\documentclass{pretty-preprint}
\usepackage{xcolor}
\usepackage[section]{placeins} 
\usepackage{wrapfig}
\usepackage[numbers,sort&compress]{natbib}
\usepackage[table]{xcolor}

\definecolor{opensourcebg}{RGB}{230,245,234}
\definecolor{closedsourcebg}{RGB}{220,220,220}

\title{RSIAgent: Autonomous Exploration for Recursive Self-improvement in New Environments}
\runningtitle{RSIAgent}

\author[1,2,\textdagger,$\ddagger$]{Sibo Zhu}
\author[1,3,\textdagger,$\ddagger$]{Shicheng Fan}
\author[1,2,\textdagger,$\ddagger$]{Xinyue Wang}
\author[1,2,$\ddagger$]{Wenyi Wu}
\author[1,*]{Kun Zhou}
\author[1]{Biwei Huang}
\affiliation[1]{Aether AI}
\affiliation[2]{University of California San Diego}
\affiliation[3]{University of Illinois Chicago}
\contribution[*]{Corresponding author and project leader}
\contribution[\textdagger]{Equal contribution}
\contribution[$\ddagger$]{Work done during internship in Aether AI}

\correspondence{Kun Zhou (\email{franciskunzhou@gmail.com})}
\metadata[Code]{\href{https://github.com/AetherLabsAI/RSIAgent}{github.com/AetherLabsAI/RSIAgent}}
\metadata[Website]{\href{https://aetherlabsai.github.io/RSIAgent/}{aetherlabsai.github.io/RSIAgent/}}

\abstract{%
Digital agents must often adapt to new environments whose interfaces, tools, and failure modes are not fully captured by pretrained models. We introduce \textbf{RSIAgent}, a training-free multi-agent framework for recursive self-improvement through autonomous memory construction. RSIAgent coordinates curriculum, actor, and verifier agents to continually explore the environment, validate outcomes, and retain environment-specific knowledge, including reusable causal relationships between actions, conditions, and consequences. It further adopts a \textbf{broad-then-deep} exploration strategy, combining parallel broad recursive self-exploration for discovering diverse environment structures with focused deep self-exploration for uncovering hard cases, hidden constraints, boundary conditions, and previously unknown causal dependencies. The resulting memory is frozen and can be directly reused for downstream tasks without updating model parameters. Experiments on OSWorld-v2 and Agent's Last Exam show that RSIAgent substantially improves strong open-source models, enabling Kimi-K3 and GLM-5.3 to outperform frontier closed-source models including GPT-6.
}


\begin{document}
\maketitle

\begin{figure}[!htbp]
  \centering
  \includegraphics[width=\linewidth]{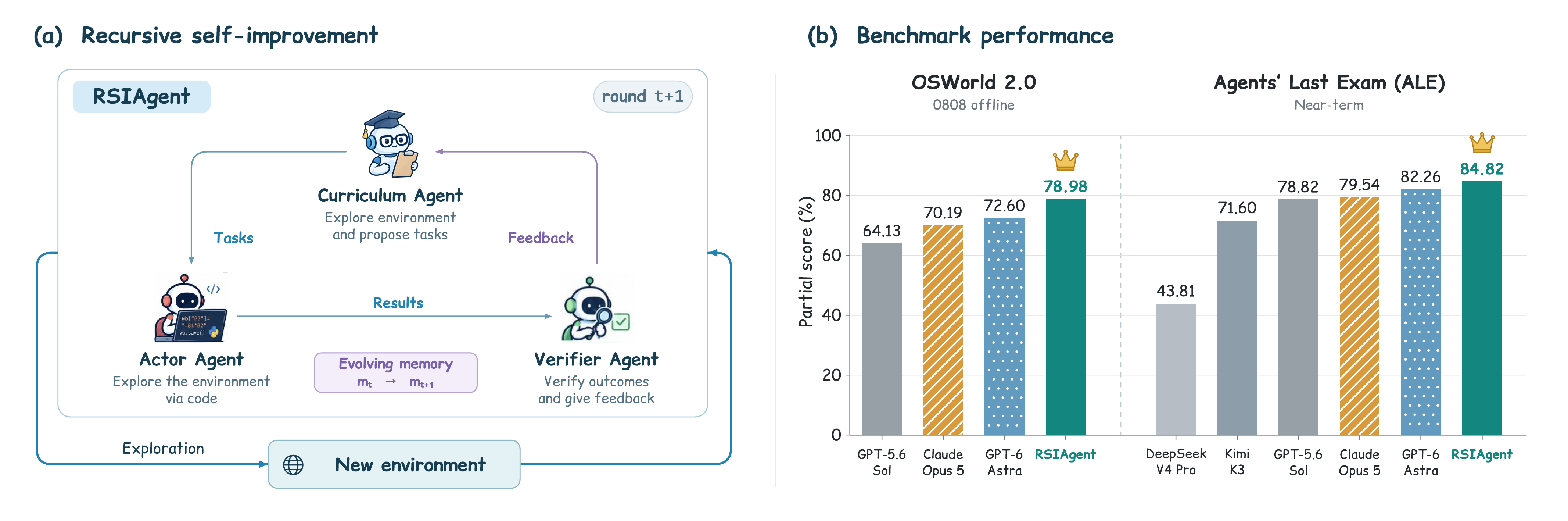}
  \caption{Overview of RSIAgent. (a) Without any gold labels or human supervision, RSIAgent autonomously explores and adapts to the new environment via the recursive curriculum-action-verification loop. (b) RSIAgent enables open-source models (Kimi-K3 and GLM-5.3) to surpass frontier closed-source models such as GPT-6 Astra, on the OSWorld 2.0 (0808 offline) and Agents' Last Exam (Near-term).}
  \label{fig:rsi-overview}
\end{figure}

\section{Introduction}
\label{sec:introduction}

Driven by scaling laws, large language models (LLMs) and vision-language models (VLMs) have demonstrated stronger abilities in perception, reasoning, planning, and tool use~\cite{zhao2026surveylargelanguagemodels}. Building on these advances, digital agent systems have emerged as a promising paradigm for automating complex user tasks in digital computer environments~\cite{xie2024osworldbenchmarkingmultimodalagents,osworld_verified,hu2025osagentssurveymllmbased}. They interact with environments by observing visual or interface information, and executing actions such as clicking or generating programs~\cite{tan2024cradleempoweringfoundationagents,wu2024oscopilotgeneralistcomputeragents,agashe2025agents2compositionalgeneralistspecialist,han2026vlaaguiknowingstoprecover,cheng2024seeclickharnessingguigrounding,zheng2024gpt4visiongeneralistwebagent,he2024webvoyagerbuildingendtoendweb}.


However, in real-world applications, digital agent systems are often required to operate in new environments whose interfaces, tools, conventions, and failure modes may not be fully captured by their pretrained knowledge. Existing adaptation approaches commonly rely on collecting additional interaction data for further training, often with human assistance. While effective, this paradigm introduces substantial cost and is difficult to apply in private or continuously changing environments~\cite{wang2026opencua,sun2026learning,hu2025osagentssurveymllmbased}. In contrast, training-free adaptation through context management offers a more flexible alternative, and recent work has shown that effectively organizing memory information within the context can substantially improve agent performance~\cite{shinn2023reflexion,zhang2026agentic}. Beyond simply memorizing successful trajectories, effective adaptation should also enable agents to discover stable causal relationships between actions, environment conditions, and outcomes, and organize these relationships into reusable causal structures. This raises a fundamental question: \emph{can an agent autonomously discover causal relations from a new environment into a reusable memory to improve itself?}


Human learning of new software often follows a recursive loop: first identifying what needs to be learned, then interacting with the environment to collect experience, and finally distilling useful knowledge from observed outcomes. Importantly, this process is not merely experience accumulation, but also a form of causal discovery: by actively trying different actions and observing their consequences, humans gradually infer which factors determine success, failure, and state transitions. Inspired by this process, we design a recursive self-improvement~(RSI) framework for digital agents in new environments. To instantiate this loop, we introduce a multi-agent framework with three complementary roles that continually expand and refine memory with newly acquired causal knowledge. The curriculum agent decides what to explore next, the actor agent interacts with the environment and updates the memory, and the verifier agent grounds observed outcomes with environment feedback, allowing the system to progressively uncover and consolidate reusable causal structures.

Building on this multi-agent framework, we devise RSIAgent, a two-stage training-free evolving strategy that autonomously explores a new environment to construct reusable memory. We adopt a \textbf{broad-then-deep} strategy: \emph{Broad Recursive Self-exploration}~(BRS) explores diverse directions in parallel to acquire an overall understanding of the environment, while \emph{Deep Recursive Self-exploration}~(DRS) focuses on important directions to uncover corner cases, hidden constraints, and boundary conditions. This coarse-to-fine process first builds broad coverage and then refines critical details, resembling the pretraining-then-posttraining paradigm in modern LLM development. The resulting memory is finally frozen and directly reused to support downstream task execution.

Empirically, RSIAgent enables strong recursive self-improvement across different open-source models. On both OSWorld-v2~\cite{yuan2026osworld20benchmarkingcomputer} and Agent's Last Exam~\cite{sun2026agentsexam}, RSIAgent substantially improves Kimi-K3 and GLM-5.3 through autonomous exploration and memory reuse, allowing these open-source models to outperform frontier closed-source models, including Claude Opus 5 and GPT-6. These results demonstrate that effective agent-level self-improvement can significantly narrow, and even reverse, the capability gap between open- and closed-source foundation models without updating model parameters.

Our main contributions are:
\begin{itemize}
    \item We propose RSIAgent, a general multi-agent framework for recursive self-improvement, enabling to autonomously acquire, verify, and reuse knowledge in new environments without updating model parameters.

    \item We design Broad Recursive Self-exploration and Deep Recursive Self-exploration to first acquire diverse environment knowledge and then refine hard cases, hidden constraints, and boundary conditions.

    \item We show that RSIAgent can recursively improve open-source models, enabling Kimi-K3 and GLM-5.3 to outperform frontier closed-source models on OSWorld-v2 and Agent's Last Exam.
\end{itemize}



\section{Preliminary}
\label{sec:setting}
In this paper, we formalize the digital agent and define the problem studied in this work.

\paragraph{Agent in Digital Environments.}
\label{sec:computer-use}


Given a task instruction $q$, a digital agent interacts with an environment $\mathcal{E}$ through a sequence of actions $a_1,\ldots,a_T$ until the task requirements are satisfied. At step $t$, the agent $\pi_\theta$ generates an executable action $a_t$ conditioned on the durable memory $M$ and the interaction history $h_t=(o_0,a_1,\ldots,a_{t-1},o_{t-1})$, where $o_0$ denotes the initial environment observation. After executing $a_t$, the environment transitions to a new state and returns a new observation:
\begin{equation}
\label{eq:digital-agent-interaction}
a_t \sim \pi_\theta(\cdot \mid q, h_t, M), \qquad
(s_t,o_t) \sim \mathcal{E}(\cdot \mid s_{t-1}, a_t).
\end{equation}
Here, $s_t$ denotes the underlying environment state, which may not be directly observable to the agent. The resulting observation $o_t$ is appended to the interaction history to form $h_{t+1}$ for the next decision. The interaction history is reset when switching to a new task, whereas the durable memory $M$ persists and retains reusable knowledge acquired across tasks.

In our implementation, we adopt a code-as-policy formulation, where each action $a_t$ is represented as an executable program. This action space provides a general interface for controlling heterogeneous software systems, including GUI applications. Our RSI framework builds on this paradigm by introducing autonomous exploration and persistent memory updates, enabling the agent to continually acquire and reuse environment-specific knowledge without modifying model parameters.




\paragraph{Problem Statement.}
\label{sec:adaptation}
In this paper, we study recursive self-improvement of an agent system in a new environment without updating model parameters. Given a new environment $\mathcal{E}$, the agent $\pi_{\theta}$ first performs autonomous exploration to construct a persistent memory $M$ that captures reusable environment-specific knowledge, and then directly reuses this memory to support downstream tasks at test time. Accordingly, the central problem is to design an exploration strategy that can efficiently identify informative experiences, ground them with reliable environment feedback, and continuously consolidate the resulting knowledge into memory. 

\section{Methodology}
\label{sec:method}

\begin{figure}[!t]
  \centering
  \includegraphics[width=\linewidth]{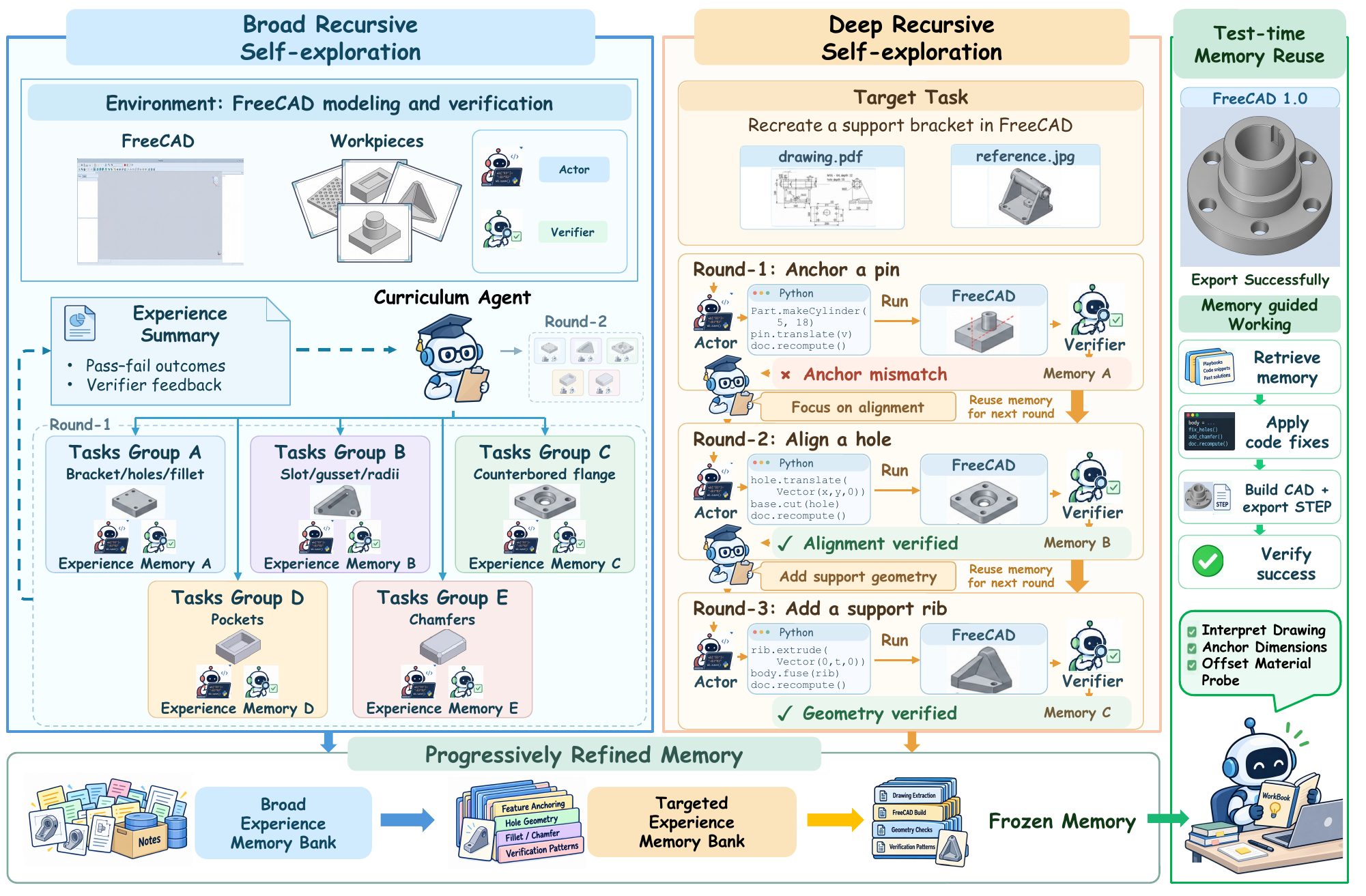}
  \caption{Overview of RSIAgent, illustrated with a FreeCAD task. BRS acquires diverse experience through parallel exploration; DRS iteratively refines memory through going deeper. The curriculum agent proposes tasks, the actor agent executes code-based actions and updates memory, and the verifier agent checks outcomes. The accumulated memory is frozen and reused for downstream tasks.}
  \label{fig:rsi-framework}
\end{figure}

In this section, we present \textbf{RSIAgent}, a recursive self-improvement framework that enables agents to autonomously explore and adapt to new environments. Section~\ref{sec:agent-harness} introduces the multi-agent harness framework, which decomposes the agent system into three collaborative agent roles. Section~\ref{sec:exploration} then presents our broad-then-deep recursive self-exploration strategy for continually updating the memory and reusing it at test time. Figure~\ref{fig:rsi-framework} provides an overview of our method.


\subsection{Unified State and Structured Workflow}
\label{sec:state-workflow}
Reliable recursive self-improvement requires a clear representation of task progress and a consistent process for updating it. StructAgent~\cite{wu2026structagent} addresses this through two complementary components: a unified state that records task requirements, useful execution information, and verified evidence; and a structured workflow that grounds progress updates in verification.

Building on this design, RSIAgent connects curriculum generation, execution, verification, and memory consolidation in a recursive workflow. The curriculum agent guides exploration with the target query as a reference, while the actor agent executes the proposed tasks and the verifier agent grounds their outcomes in environment feedback. The resulting experience is consolidated into persistent memory and informs subsequent curriculum decisions and execution.

\subsection{Multi-Agent Harness Framework}
\label{sec:agent-harness}
To better control autonomous exploration and self-improvement in a new environment, we design a multi-agent harness framework that can operate through a coordinated recursive loop. The framework consists of an actor agent with evolvable memory, a verifier agent for environment feedback grounding, and a curriculum agent for guiding exploration.


\paragraph{Actor Agent with Evolvable Memory.}
\label{sec:memory}
The actor agent is the primary policy model in the multi-agent system, responsible for understanding the environment and generating executable actions. To support adaptation across downstream tasks in an environment, the actor agent is equipped with a persistent memory that stores environment-specific knowledge, reusable procedures and scripts, and lessons learned from previous executions. This memory is evolvable: after the verifier agent evaluates an outcome, the actor agent consolidates the grounded experience by adding new knowledge and revising or removing outdated information when necessary. The updated memory is then inherited by subsequent actor agent instances, allowing useful knowledge to accumulate throughout exploration.

\paragraph{Verifier Agent with Environment Feedback.}
The verifier agent serves as the independent evaluator in the multi-agent system, responsible for determining whether the actor agent's execution has successfully satisfied the task requirements. To make this judgment reliable, it directly inspects the feedback from the environment, including execution results, interface states, and other observable evidence, and grounds its decision in these signals. The verifier agent is isolated from the actor agent's private reasoning and memory, which helps reduce correlated errors during evaluation. Based on grounded evidence, it returns a success or failure judgment together with supporting feedback, which is then used to determine whether the corresponding experience should be consolidated into the memory.

\paragraph{Curriculum Agent for Guiding Exploration.}
The curriculum agent serves as the high-level coordinator in the multi-agent system, responsible for deciding what the system should explore next. Concretely, it generates suitable practice tasks for the actor and verifier agents based on the current target, accumulated memory, and previous exploration outcomes. By selecting prerequisite skills, informative variants, failure-driven practice, and stress-test cases, the curriculum agent determines the direction of exploration and progressively expands the coverage of the evolvable memory. Exploration continues until the generated tasks are unlikely to contribute substantial new knowledge.

\subsection{Multi-stage Autonomous Exploration for RSI}
Building on the multi-agent framework, RSIAgent organizes autonomous exploration into two complementary stages: \emph{Broad Recursive Self-exploration}~(BRS), which discovers diverse and reusable experiences to build a broad understanding of the environment, and \emph{Deep Recursive Self-exploration}~(DRS), which refines the accumulated knowledge through target-driven practice. 
After exploration, the resulting memory is frozen and directly reused for final evaluation.

\paragraph{Stage-1: Broad Recursive Self-exploration.}
\label{sec:exploration}
Broad Recursive Self-exploration (BRS) aims to rapidly build a broad understanding of a new environment by collecting diverse interaction experiences. BRS follows a recursive exploration loop: the curriculum agent first generates exploration tasks, the actor agents execute them, and the verifier agents evaluate the resulting outcomes. To ensure broad coverage, at each iteration, the curriculum agent proposes multiple tasks spanning different exploration directions, which are executed and verified in parallel. After collecting the resulting trajectories and verified feedback, the curriculum agent uses the accumulated experience to identify remaining knowledge gaps and generate more informative tasks for the next iteration. Through this recursive process, BRS progressively expands the coverage of environment-specific knowledge, reusable procedures, and failure patterns.

\paragraph{Stage-2: Deep Recursive Self-exploration.}
\label{sec:practice}
Deep Recursive Self-exploration (DRS) aims to refine the accumulated memory by focusing on important knowledge gaps, hard cases, and boundary conditions revealed during task execution. DRS follows a sequential recursive loop that progressively increases exploration difficulty. The curriculum agent first proposes a challenging task that is likely to expose unpredictable issues, hidden constraints, or weaknesses in the current memory. The actor agent then attempts the task using the accumulated memory, while the verifier agent evaluates the outcome and provides grounded feedback. Based on the resulting successes, failures, and newly revealed uncertainties, the curriculum agent generates a more challenging follow-up task for the next iteration. Each verified experience is consolidated into memory before the subsequent task is proposed, allowing DRS to continuously push the agent toward harder and less explored cases.

\paragraph{Test-time Memory Reuse.}
\label{sec:evaluation}
After exploration, the accumulated memory is frozen and provided to the actor agent for test-time use. At this stage, the curriculum agent and all memory updates are disabled. Given a target task, the actor agent directly reuses the procedures, discovered constraints, and failure lessons stored in memory to guide its actions, while the verifier agent evaluates the resulting outcome against the task requirements. This action--verification loop continues until the verifier agent confirms that all task requirements have been satisfied.


\section{Experiments}
\label{sec:experiments}


\subsection{Experimental Setup}
\label{sec:experimental-setup}

\paragraph{Benchmarks and Metrics.}
We evaluate RSIAgent on OSWorld 2.0 (0808 offline) and Agents'
Last Exam (ALE) Near-term, which require agents to complete tasks
in interactive software environments
\cite{yuan2026osworld20benchmarkingcomputer,sun2026agentsexam}.
Our OSWorld results are aggregated over 82 offline tasks and ALE results cover all 67 Near-term tasks.
We report partial score, the mean task score, and binary accuracy, the
proportion of tasks receiving full credit. Both metrics are
expressed as percentages. Detailed benchmark descriptions, agent
configurations, and RSI task-selection and reporting protocols are
provided in Appendix~\ref{app:experimental-setup}.

\paragraph{Implementation Details.}
Our configuration uses a shared code-as-policy harness for RSIAgent. GLM-5.3 serves as the default actor agent, while Kimi-K3 serves as the verifier agent and the curriculum agent in a separate context. Both exploration stages use the target query as a reference for the curriculum agent. For Broad Recursive Self-exploration, we set a nominal budget of eight exploration projects, with up to four projects executed concurrently. The budget is checked between completed waves, without interrupting an ongoing wave. For Deep Recursive Self-exploration, exploration proceeds sequentially until the curriculum agent determines that no further useful practice is needed. In this stage, a successful practice does not automatically terminate exploration: the curriculum agent reviews the verified outcome and accumulated memory to decide whether additional practice is worthwhile. After exploration, the accumulated memory is frozen and reused for evaluation.


\begin{table}
  \centering
  \caption{Model comparison on OSWorld 2.0 (0808 offline) and
  Agents' Last Exam Near-term. Baseline results are mostly copied from their official technical reports or blogs.}
  \label{tab:main-results}
  \renewcommand{\baselinestretch}{1}
  \fontsize{9.5}{11.5}\selectfont
  \setlength{\tabcolsep}{4.5pt}
  \renewcommand{\arraystretch}{1.18}
  \begin{tabularx}{\linewidth}{@{}
    >{\raggedright\arraybackslash}p{0.30\linewidth}
    *{2}{>{\centering\arraybackslash}X}
    @{\hspace{1.2em}}
    *{2}{>{\centering\arraybackslash}X}@{}}
    \toprule[0.7pt]
    & \multicolumn{2}{c}{\shortstack{\textbf{OSWorld 2.0}\\[1pt]
        {\fontsize{8.2}{10}\selectfont 0808 offline / 82 tasks}}}
      & \multicolumn{2}{c}{\shortstack{\textbf{Agents' Last Exam}\\[1pt]
        {\fontsize{8.2}{10}\selectfont Near-term / 67 tasks}}} \\
    \cmidrule(lr){2-3}\cmidrule(l){4-5}
    \textbf{Model / method}
      & \textbf{Partial (\%)} & \textbf{Binary (\%)}
      & \textbf{Partial (\%)} & \textbf{Binary (\%)} \\
    \midrule[0.35pt]
    \rowcolor{opensourcebg}
    \multicolumn{5}{@{}l}{\textit{Open-source Models}} \\
Kimi-K2.6 & 22.10 & 4.60 & 21.70 & 9.20 \\
MiMo-V2.5 & --- & --- & 23.60 & 8.60 \\
    DeepSeek V4 Pro & --- & --- & 43.81 & 19.90 \\
    Qwen3.8-Max & --- & --- & 52.50 & 27.00 \\
    Kimi-K3 & 58.30 & --- & 71.60 & 40.30 \\
    \midrule[0.35pt]
    \rowcolor{closedsourcebg}
    \multicolumn{5}{@{}l}{\textit{Closed-source Models}} \\
    Claude Opus 4.8 & 54.80 & 20.60 & 64.00 & 43.30 \\
    Gemini-3.8-Flash & 59.00 & --- & --- & --- \\
    GPT-5.6 Sol & 64.13 & 28.10 & 78.82 & 47.76 \\
    Muse Spark 1.3 & 66.90 & --- & --- & --- \\
    Claude Fable 5 & --- & --- & 71.10 & 37.30 \\
    Claude Opus 5 & 70.19 & 34.72 & 79.54 & 46.27 \\
    GPT-6 Astra & 72.60 & --- & 82.26 & 52.24 \\
    \midrule[0.35pt]
    \rowcolor{opensourcebg}
    \multicolumn{5}{@{}l}{\textit{Ours (Using Open-source Models)}} \\
    RSIAgent (w/o RSI) & 71.97 & 37.80 & 83.75 & 49.25 \\
    \textbf{RSIAgent} & \textbf{78.98} & \textbf{42.68}
      & \textbf{84.82} & \textbf{50.75} \\
    \bottomrule[0.7pt]
  \end{tabularx}
  \par\vspace{4pt}
  {\fontsize{8.2}{10.2}\selectfont\raggedright
    ---: not reported in the cited benchmark leaderboards or its official technical reports.\par}
\end{table}

\subsection{Main Results}
\label{sec:main-results}

\paragraph{Effect of Recursive Self-Improvement.}
RSI improves the reported aggregate performance of the existing agent harness on both benchmarks. As shown in Table~\ref{tab:main-results}, OSWorld partial score increases from 71.97 to 78.98, while binary accuracy increases from 37.80 to 42.68. On ALE, Partial increases from 83.75 to 84.82, and binary accuracy increases from 49.25 to 50.75. Two-stage evolving brings clear gains extending both procedure correctness and full task completion. For the tasks without a completed RSI result, we retain the baseline scores for them (see Appendix~\ref{app:rsi-task-selection} for task-selection and aggregation details).

\paragraph{Comparison with Frontier Models.}
RSIAgent achieves the highest reported partial-credit scores among the systems compared in Table~\ref{tab:main-results}. Using GLM-5.3 and Kimi-K3, it reaches 78.98 on OSWorld 2.0 and 84.82 on ALE, exceeding the reported GPT-6 Astra scores by 6.38 and 2.56 percentage points, respectively, and also scoring above Claude Opus 5 on both benchmarks.
These results highlight the value of combining our multi-agent harness framework with two-stage exploration strategy and memory reuse to strengthen open-source models without updating model parameters.

\subsection{Effect of RSI Rounds}
\label{sec:rsi-rounds}

\begingroup
\setlength{\intextsep}{6pt}
\setlength{\columnsep}{12pt}
\begin{wrapfigure}{r}{0.58\textwidth}
  \centering
  \includegraphics[width=\linewidth]{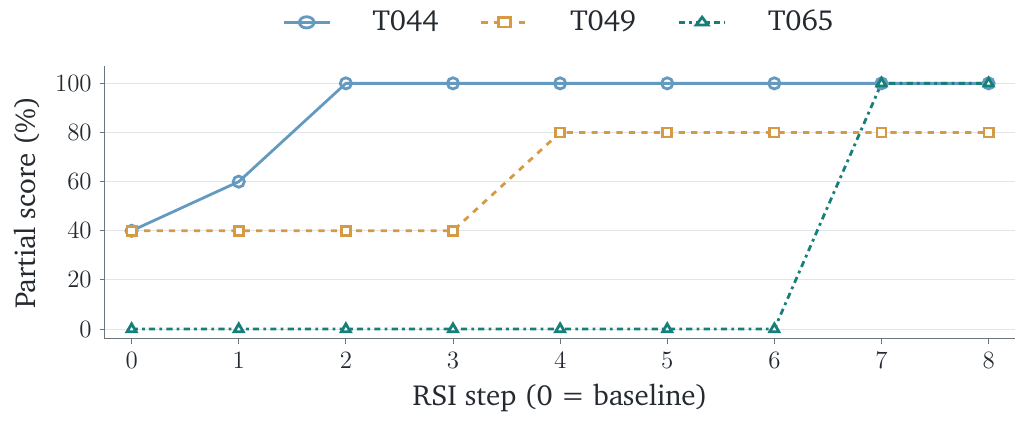}
  \captionsetup{font=footnotesize}
  \caption{Partial scores across RSI steps for three OSWorld 2.0 tasks.
  Curves show T044, T049, and T065, with RSIAgent (w/o RSI)
  as the baseline.}
  \label{fig:rsi-rounds}
\end{wrapfigure}
We examine RSI performance on three representative OSWorld 2.0 tasks:
T044 (video editing), T049 (presentation repair), and T065 (railway
booking). Figure~\ref{fig:rsi-rounds} presents their partial-score
profiles, with RSIAgent (w/o RSI) as the baseline. The horizontal axis
shows steps 0--8, with step 0 denoting the baseline. By step 8,
T044, T049, and T065 reach scores of
100\%, 80\%, and 100\%, respectively. BRS progressively
accumulates diverse procedures and environment knowledge in memory,
while DRS refines task-specific details. As this knowledge comes to
cover a target task's critical requirements, resolving a remaining
bottleneck can produce a discrete score increase. Such single-task
evaluations reveal these breakthroughs more readily than the gradual
memory accumulation that precedes them.
Detailed case studies of how memory grows and supports task execution
are provided in Appendix~\ref{app:memory-case-studies}.
\par
\ifnum\value{WF@wrappedlines}>1\relax
  \vspace{\dimexpr\value{WF@wrappedlines}\baselineskip-\baselineskip\relax}
\fi
\WFclear
\endgroup
\FloatBarrier

\subsection{Ablation Study}
\label{sec:ablations}

\begin{figure}[!ht]
  \centering
  \includegraphics[width=\linewidth]{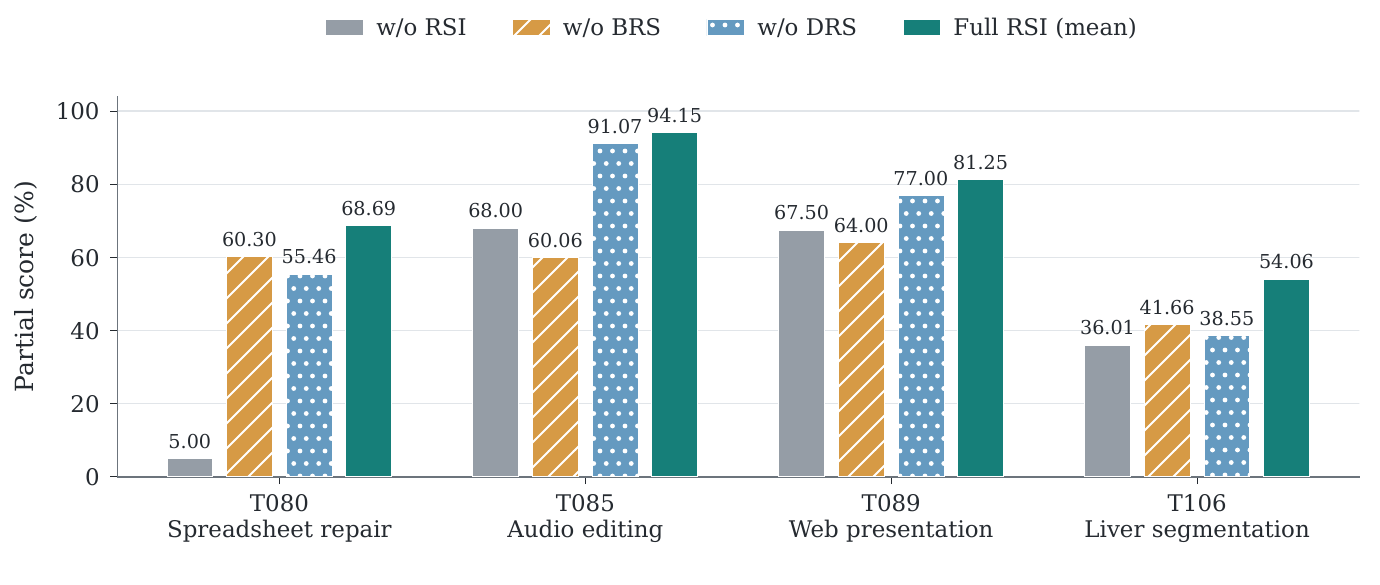}
  \caption{Stage comparison on four OSWorld 2.0 tasks. Full RSI bars
  average two historical evaluations.}
  \label{fig:ablations}
\end{figure}

We compare the exploration stages on four OSWorld 2.0 (0808 offline)
tasks: T080 (WPS spreadsheet repair), T085 (REAPER audio editing),
T089 (browser-based presentation repair), and T106 (3D Slicer liver
segmentation). The w/o BRS variant skips
\emph{Broad Recursive Self-exploration} and performs only
\emph{Deep Recursive Self-exploration} from empty memory. Conversely,
w/o DRS evaluates the memory acquired through broad exploration alone,
while w/o RSI directly evaluates the agent with empty memory. Full RSI
combines both stages. Figure~\ref{fig:ablations} reports task-level
partial scores; task selection, repetition counts, and exploration
budgets are detailed in Appendix~\ref{app:ablation-protocol}.

Combining broad and deep exploration yields the highest reported partial
score on all four tasks. Full RSI reaches a mean score of 74.54\%, compared
with 65.52\% for broad-only exploration and 56.50\% for deep-only
exploration. Broad-only exploration improves over the baseline on
every task, whereas deep-only exploration falls below the baseline
on T085 and T089. These observations support combining
\emph{Broad Recursive Self-exploration} with subsequent
\emph{Deep Recursive Self-exploration}.

\FloatBarrier

\subsection{Evaluation in Game Environments}
\label{sec:game-evaluation}

We further evaluate whether RSIAgent generalizes beyond standard computer-use tasks to autonomous game development, an interactive setting that requires agents to repeatedly play, diagnose, and modify executable games. We randomly sample 40 tasks from GameCraft-Bench~\citep{luo2026gamecraft} and compare RSIAgent against Play2Code~\citep{huang2026gui}, a continual game-improvement baseline based on iterative playtesting and code revision. Detailed model configurations and development budgets are provided in Appendix~\ref{app:game}.

As shown in Table~\ref{tab:game-results}, \textbf{first}, RSIAgent substantially improves game quality across all base generators. \textbf{Second}, while Play2Code improves games from weaker generators, its benefit diminishes as the base game becomes stronger and it can even degrade already high-quality games. In contrast, RSIAgent consistently improves both weak and strong base games. \textbf{Third}, incorporating RSI experience further improves quality. The accumulated experience provides reusable knowledge for diagnosing and editing different types of games.

\begin{table}[h]
\centering
\caption{
Performance on 40 GameCraft-Bench tasks.
Rows are grouped by the generator of the frozen base game $P_0$.
Within each group, all development methods use the same starting game
and development backbone. The best score for each quality metric
within each group is shown in \textbf{bold}.
}
\label{tab:game-results}
\small

\begin{tabular*}{\textwidth}{@{\extracolsep{\fill}}lccccc}
\toprule
Method
& Mechanics
& Depth
& Visuals
& Art
& Overall $\uparrow$ \\
\midrule

\multicolumn{6}{l}{\emph{Generator: Codex + GPT-5.5 (high)}} \\
Baseline
& 61.5 & 53.0 & 54.7 & 47.9 & 52.77 \\
\quad + Play2Code
& 60.0 & 50.2 & 52.2 & 47.5 & 51.05 \\
\quad + RSIAgent (w/o RSI)
& 66.4 & 57.1 & 59.3 & 53.8 & 57.84 \\
\quad + RSIAgent
& \textbf{69.7} & \textbf{61.0} & \textbf{62.6} & \textbf{57.1}
& \textbf{61.28} \\

\midrule
\multicolumn{6}{l}{\emph{Generator: Kimi-K2.6}} \\
Baseline
& 43.5 & 33.5 & 34.1 & 22.6 & 31.28 \\
\quad + Play2Code
& 48.9 & 36.5 & 40.9 & 27.9 & 36.02 \\
\quad + RSIAgent (w/o RSI)
& 55.2 & 43.1 & 47.0 & 34.6 & 42.61 \\
\quad + RSIAgent
& \textbf{59.0} & \textbf{47.2} & \textbf{50.8} & \textbf{38.1}
& \textbf{46.37} \\

\midrule
\multicolumn{6}{l}{\emph{Generator: GLM-5.3-Flash}} \\
Baseline
& 36.5 & 29.6 & 31.4 & 28.5 & 30.55 \\
\quad + Play2Code
& 48.6 & 37.6 & 40.0 & 33.7 & 38.25 \\
\quad + RSIAgent (w/o RSI)
& 55.1 & 44.0 & 47.3 & 40.1 & 44.73 \\
\quad + RSIAgent
& \textbf{59.2} & \textbf{48.1} & \textbf{51.3} & \textbf{43.8}
& \textbf{48.72} \\

\midrule
\multicolumn{6}{l}{\emph{Generator: Qwen3.8-27B}} \\
Baseline
& 50.5 & 36.4 & 39.7 & 42.8 & 41.30 \\
\quad + Play2Code
& 58.1 & 43.4 & 45.3 & 48.4 & 47.67 \\
\quad + RSIAgent (w/o RSI)
& 64.3 & 50.2 & 52.1 & 55.0 & 53.82 \\
\quad + RSIAgent
& \textbf{68.0} & \textbf{54.1} & \textbf{55.9} & \textbf{58.7}
& \textbf{57.46} \\

\bottomrule
\end{tabular*}
\end{table}

\subsection{Failure Modes Analysis}
\label{sec:failure-analysis}

\begin{figure}[htbp]
  \centering
  \includegraphics[width=\linewidth]{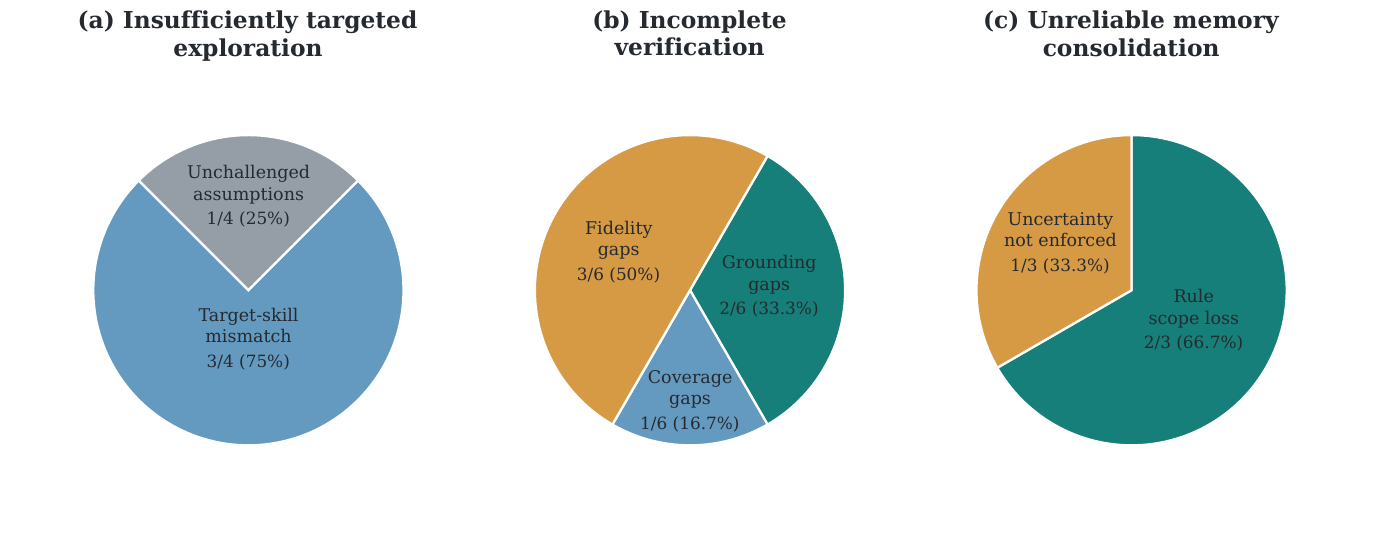}
  \caption{Failure-mode subtypes from selected case audits. Labels show
  counts and within-mode percentages;
  cases may contribute to multiple panels.}
  \label{fig:failure-mechanisms}
\end{figure}

Our failure mode analysis identifies three mechanisms that limit
recursive self-improvement: insufficiently targeted exploration,
incomplete verification, and unreliable memory consolidation.
These mechanisms can interact, allowing an initially uncertain or potentially incorrect operations, to persist through practice and influence subsequent agent behaviors. Figure~\ref{fig:failure-mechanisms} summarizes these mechanisms.

\paragraph{Insufficiently Targeted Exploration.}
Additional practice may leave target-specific weaknesses unresolved
when it does not challenge the decisions responsible for them.
In the inspected trajectories, the curriculum agent generated
practice involving document processing, conflicting information,
and submission persistence, while obtaining unavailable user
information remained untested. The problematic missing-information
rules consequently remained in memory. These observations highlight
the importance of selecting experiences that challenge uncertain
knowledge and address unresolved target requirements, beyond
expanding the diversity of practice tasks.

\paragraph{Incomplete Verification.}
Local verification can approve an execution without establishing
that all task requirements have been satisfied. In the audited
cases, the verifier agent returned \texttt{PASS} despite unsupported
field values or artifact discrepancies under the official rubric.
Checking that an intended output was produced does not establish
that the underlying interpretation or requirement checklist was
complete. This gap can affect both final execution and subsequent
learning, because an accepted mistake may become the basis for
future memory updates and exploration decisions.

\paragraph{Unreliable Memory Consolidation.}
An inadequately verified decision can become a reusable rule that
negatively influences subsequent execution. In the inspected
form-completion runs, the actor agent retained rules that treated
missing-data markers as valid answers or interpreted unavailable
information as a negative answer. Subsequent attempts reused these
rules without obtaining the missing facts. This failure illustrates
that useful memory must preserve the conditions and uncertainty
of an experience, rather than treating local acceptance as evidence
of general validity.






\section{Related Work}
\label{sec:related-work}

\paragraph{Digital Agents.} Digital agents use language-model reasoning and tool use to complete tasks in software environments~\citep{hu2025osagentssurveymllmbased,zhao2026surveylargelanguagemodels}. Benchmarks evaluate their capabilities across desktop applications and tool environments~\citep{sun2026agentsexam,xi2026toolgym,xie2024osworldbenchmarkingmultimodalagents,yuan2026osworld20benchmarkingcomputer}, including changing user requirements and the quality of execution outcomes~\citep{fan2026agentic,huang2026gui,luo2026gamecraft,miao2025recode,zou2026users}. Research has improved visual grounding and GUI interaction~\citep{cheng2024seeclickharnessingguigrounding,he2024webvoyagerbuildingendtoendweb,zheng2024gpt4visiongeneralistwebagent}. OpenCUA~\citep{wang2026opencua} learns from human demonstrations, UI-TARS-2~\citep{wang2025ui} uses multi-turn reinforcement learning, and EvoCUA~\citep{xue2026evocua} combines synthetic experience with policy optimization. CodeAct represents actions as executable programs that can be revised using execution feedback~\citep{wang2024executable}. Cradle, OS-Copilot, and Agent S2 support application control, reusable skills, and coordinated execution~\citep{tan2024cradleempoweringfoundationagents,wu2024oscopilotgeneralistcomputeragents,agashe2025agents2compositionalgeneralistspecialist}. Other studies improve information organization, planning, and failure recovery~\citep{han2026vlaaguiknowingstoprecover,jiang2023structgpt,sun2026learning,wu2026planner,wu2026structagent}, while research on external evidence and feedback incentives examines ways to improve reliability~\citep{fan2026verifiable,min2026quco,yang2026paying}. We focus on adapting pretrained agents to new software environments through autonomous exploration. RSIAgent uses code-as-policy to acquire the environment-specific knowledge needed for downstream tasks, without further model training.

\paragraph{Recursive Self-Improvement of Agents.} Research on agent self-improvement explores changes to agent designs~\citep{hu2025automated,razzhigaev2026ouroboros,robeyns2025selfimprovingcodingagent}, prompts and skills~\citep{agrawal2026gepa,yang2026skillopt}, and improvement procedures~\citep{wang2026metaskill,zelikman2024selftaughtoptimizerstoprecursively}. DGM evaluates self-modifications on coding benchmarks~\citep{zhang2026darwingodelmachineopenended}, while HGM~\citep{wang2026huxley} uses descendant performance to identify agents with greater potential for further improvement. Hyperagents~\citep{zhang2026hyperagents} allows a meta-agent to modify both itself and the task agent, enabling the improvement procedure to evolve alongside task-solving capabilities. These methods study self-improvement with access to benchmark scores or labeled reference data. Agents can also learn from generated training experience~\citep{lu2026arex,xia2025agent0,zhangscaling} or accumulate experience in reflections, skills, manuals, and persistent memory~\citep{chen2024automanual,karten2026prime,shinn2023reflexion,wang2023voyager}. Recent work improves how this knowledge is organized, updated, and retrieved~\citep{wu2025auto,wu2026towards,wu2026gam,yang2026selfmem,zhang2026memskill,zhang2026agentic}. HyMEM combines symbolic nodes and trajectory embeddings in a hybrid memory graph, supporting multi-hop retrieval and inference-time memory updates for computer-use agents~\citep{zhu-etal-2026-hybrid}. Voyager uses an automatic curriculum to acquire skills~\citep{wang2023voyager}. EchoTrail-GUI and ZhuLong explore environments to collect reusable experience~\citep{li2026echotrail,liu2026zhulong}, while CoEvoSkills and Recuris use verification feedback to refine skills and memory~\citep{zhang2026coevoskills,yu2026recursive}. We study recursive self-improvement in new digital environments without external task rewards. RSIAgent autonomously explores the environment under a broad-then-deep exploration strategy and uses environment feedback to construct reusable memory. This memory guides further exploration and supports downstream task execution without updating model parameters.

\section{Conclusion}
\label{sec:conclusion}
In this work, we introduce \textbf{RSIAgent}, a training-free multi-agent framework for recursive self-improvement in new digital environments. By coordinating curriculum, actor, and verifier agents, RSIAgent autonomously acquires, verifies, and consolidates environment-specific knowledge into reusable memory. Its broad-then-deep exploration strategy first builds diverse environment coverage through broad recursive self-exploration~(BRS), and then focuses on hard cases, hidden constraints, and boundary conditions through deep recursive self-exploration~(DRS). Experiments on OSWorld-v2 and Agent's Last Exam show that RSIAgent can substantially improve strong open-source models and enable them to outperform frontier closed-source models without updating model parameters.

In future work, we plan to extend recursive self-improvement beyond digital computer-use environments to broader interactive domains, including AI for Science, where agents must learn specialized tools, workflows, and scientific procedures, and games, which provide complex and continuously evolving environments for studying long-horizon exploration, adaptation, and self-improvement.

\section{Limitations}
\label{sec:limitations}

RSIAgent has several limitations. First, recursive self-improvement requires additional test-time exploration and practice, which can introduce substantial computation cost. Second, performance depends on finite exploration budgets, stopping policies, and the quality of the learned memory. Third, the model-based verifier agent may produce incorrect judgments that propagate into later exploration and memory updates. Finally, different environments may require different tools, verification signals, and exploration strategies, and our current experiments do not fully isolate the contribution of every component. 
For ethical concerns, RSIAgent can autonomously explore software environments, execute programs, and retain reusable memory, introducing risks such as unintended actions, unauthorized access, and privacy leakage. Our experiments are conducted in controlled environments with permitted tools and data access. 

\clearpage
\phantomsection
\addcontentsline{toc}{section}{References}
\bibliographystyle{plainnat}
\bibliography{reference}

\beginappendix
\setcounter{figure}{0}
\setcounter{table}{0}
\renewcommand{\thefigure}{A\arabic{figure}}
\renewcommand{\thetable}{A\arabic{table}}

\section{Implementation and Agent Interfaces}
\label{app:implementation}

RSIAgent implements recursive improvement through explicit interfaces
for execution, verification, experience selection, and memory updates.
This appendix describes the target-conditioned reference implementation:
the target query guides exploration, while persistent memory carries
learned experience between attempts. Model configurations and execution
limits are given in Appendix~\ref{app:experimental-setup}; reporting
qualifications are retained in Appendix~\ref{app:main-table-reporting}.

\subsection{Role Interfaces and Information Boundaries}
\label{app:agent-interfaces}

\paragraph{Role-Specific Contexts.}
Separate contexts give each agent the information needed for its role
without sharing private reasoning across roles.
The actor agent receives the task and available memory, the verifier
agent inspects the candidate against the task, and the curriculum
agent uses previous outcomes and accumulated memory to select new
experiences. Table~\ref{tab:agent-interfaces} summarizes these interfaces.
The curriculum agent's context persists within an exploration lineage;
the verifier agent's context persists across candidate revisions within
a target attempt, but is not shared across distinct projects or target
attempts.

\begin{table}[htbp]
  \centering
  \caption{Agent interfaces in the reference RSI implementation.}
  \label{tab:agent-interfaces}
  \small
  \setlength{\tabcolsep}{5pt}
  \renewcommand{\arraystretch}{1.14}
  \begin{tabularx}{\linewidth}{@{}
    >{\raggedright\arraybackslash}p{0.16\linewidth}
    >{\raggedright\arraybackslash}X
    >{\raggedright\arraybackslash}X@{}}
    \toprule
    Agent & Observed inputs & Outputs and authority \\
    \midrule
    actor agent
      & Task instruction, task-visible environment, execution feedback,
        and available persistent memory
      & Executable actions and candidate artifacts; after verification,
        memory updates and a learning diagnosis \\
    verifier agent
      & Task requirements and candidate environment; actor-private
        memory, reasoning, and execution logs are hidden
      & Evidence-grounded local verdict and findings;
        no durable memory updates \\
    curriculum agent
      & Target query, completed exploration outcomes, actor learning
        diagnoses when available, and a disposable copy of current memory
      & Practice instructions, input fixtures, and continuation or
        stopping decisions; no candidate grades or canonical memory edits \\
    \bottomrule
  \end{tabularx}
\end{table}

\paragraph{Program-Based Actions.}
The actor interface exposes complete programs as the unit of action.
A \texttt{program} response specifies Python or Bash code, and the
runtime returns its combined output, exit status, and execution metadata.
The actor agent can request visual evidence through \texttt{look},
ask for missing information through \texttt{ask} when a user channel
is configured, and submit a \texttt{done} response with read-only checks.
The prompt directs the actor agent to investigate before acting,
repair programs using execution feedback, and check the task's actual
requirements. Code-based control does not relax requirements concerning
named applications, editable artifacts, rendered appearance, or workflow.

\paragraph{Curriculum Handoffs.}
Practice tasks specify the desired outcome without supplying a solution.
For BRS, the curriculum agent publishes a structured handoff containing
a \texttt{decision}, a \texttt{rationale}, and a list of projects with
unique identifiers and self-contained instructions. Each project has
its own input fixtures, replayed under a common project path in an
isolated environment. The actor agent receives the project instruction
and fixtures, not the curriculum agent's private search rationale.
For DRS, a \texttt{PROJECT} handoff similarly contains the practice
request and required inputs; alternative decisions return control to
the target attempt or record that exploration has stalled.

\paragraph{Independent Verification.}
The verifier interface grounds acceptance in the candidate rather than
the actor agent's account of its work. Its prompt requires checks
derived from the original task and supporting observations from the
environment. Candidate inspection uses reset-and-replay or checkpoint
restoration to prevent verifier probes from changing the submitted
artifact. The reference practice interface produces one grounded
\texttt{PASS} or \texttt{FAIL} per project, without a same-project repair
cycle. The target interface additionally supports \texttt{UNVERIFIED}:
the actor agent can supply evidence for independent reinspection, but
an unresolved outcome does not become a successful or failed learning
example.

\subsection{Memory Updates and Parallel Reconciliation}
\label{app:memory-protocol}

\paragraph{Experience-Owned Learning.}
The actor agent that produced an experience also decides what to retain
from it. After a grounded \texttt{PASS} or \texttt{FAIL}, its existing
context receives the complete verifier report and enters two learning
steps. Distillation identifies useful procedures, constraints, and
failure lessons; reconciliation checks the resulting memory against
older entries, revising contradictions and qualifying unsupported
conclusions. A failed project can contribute useful evidence without
being recorded as a verified success. The memory remains a collection
of actor-authored files, with no required schema, file count, or length.

\paragraph{Canonical Memory Ownership.}
Only completed actor learning updates are promoted to the persistent
memory bank. Work-phase memory edits are discarded before learning,
and the actor agent receives the current canonical memory at the
update boundary. The curriculum agent may inspect a disposable copy
to guide exploration, but its local edits are not synchronized back.
Neither the curriculum agent nor the verifier agent approves memory
wording. An actor-authored learning diagnosis communicates useful
conclusions and uncertainties to the curriculum agent without passing
the actor agent's private reasoning transcript.

\paragraph{Parallel Work, Sequential Updates.}
BRS separates concurrent experience acquisition from ordered memory
updates. Every project in a wave begins with the same immutable
pre-wave memory snapshot and cannot observe sibling work or outcomes.
Once all project verdicts are available, the original actor contexts
resume one at a time in the order authored by the curriculum agent.
Each update operates on the latest canonical memory, including
preceding updates from the same wave. The next wave is selected only
after these commits finish, allowing the curriculum agent to respond
to both verified outcomes and the knowledge actually retained.

\subsection{Sequential Refinement and Evaluation Lifecycle}
\label{app:implementation-lifecycle}

\paragraph{Target-Conditioned Refinement.}
The reference DRS implementation begins with an attempt at the target
query using the memory inherited from BRS. A grounded outcome is
followed by actor-owned learning, after which the curriculum agent
reviews the outcome, learning diagnosis, and updated memory.
It may select sequential practice to investigate a failure or test
an uncertain successful procedure. Each practice project is verified
and consolidated before the next curriculum decision; once practice
returns control, the target is attempted again in a reset environment
using the updated memory. All attempts retain the same agent framework.

\paragraph{Stopping Decisions.}
Stopping distinguishes the expected value of further practice from
the correctness of the target candidate. Under the default
\texttt{curriculum\_review} policy, a target \texttt{PASS} followed by
no additional practice can complete DRS; any additional practice
requires another target attempt. If the curriculum agent returns
\texttt{STALLED} after a failed target or further practice, the runner
performs one final target attempt and records its actual verdict.
The explicit \texttt{verifier\_pass} variant instead ends DRS after a
grounded target \texttt{PASS} and memory consolidation, without a
curriculum review. Unresolved verification and infrastructure failures
block phase advancement rather than being labeled convergence.

\paragraph{Frozen-Memory Evaluation.}
Final evaluation reuses learned memory without continuing the learning
loop. The runner records the frozen memory's file-tree hash, copies it
into the evaluation environment, and disables curriculum decisions
and host memory writeback. The actor--verifier harness then attempts
the target after an environment reset, and the official evaluator
scores the resulting candidate outside all agent contexts.
The runner checks memory integrity and records the protocol, stopping
policy, terminal status, and evaluation result separately.
Algorithm~\ref{alg:rsi-implementation} summarizes this reference lifecycle;
historical and selected-run variants retain their own protocol records.

\newcounter{rsiappendixalgorithm}
\renewcommand{\thersiappendixalgorithm}{A\arabic{rsiappendixalgorithm}}
\begin{center}
  \begin{minipage}{\linewidth}
    \small
    \refstepcounter{rsiappendixalgorithm}
    \label{alg:rsi-implementation}
    \noindent\textbf{Algorithm~\thersiappendixalgorithm.
    Target-conditioned RSI reference procedure.}\par\smallskip
    \hrule\smallskip
    \noindent\textbf{Inputs:} target query $q$, resettable environment,
    initial memory $M_0$, and declared stopping policy.\\
    \textbf{Outputs:} frozen memory, final candidate, terminal status,
    and externally recorded benchmark score.
    \begin{enumerate}
      \setlength{\itemsep}{3pt}
      \setlength{\parsep}{0pt}
      \item Initialize canonical memory $M\leftarrow M_0$.
      \item \textbf{BRS:} repeat complete waves until a recorded
        curriculum stop or the complete-wave budget boundary.
        \begin{enumerate}
          \setlength{\itemsep}{2pt}
          \setlength{\parsep}{0pt}
          \item The curriculum agent selects independent projects
            using $q$, $M$, and completed wave outcomes.
          \item Snapshot $M$. Execute projects and obtain verifier
            verdicts in parallel in isolated environments.
          \item After all verdicts are grounded, resume the same
            actor contexts in curriculum-authored order to distill
            and reconcile each experience into the current $M$.
        \end{enumerate}
      \item \textbf{DRS:} attempt $q$ after an environment reset and
        obtain a grounded target verdict. The same actor context
        consolidates that outcome into $M$.
      \item If the policy is \texttt{verifier\_pass} and the target
        passed, proceed to Step~7. Otherwise, ask the curriculum
        agent to select the next experience.
      \item Execute each selected practice project sequentially:
        actor execution, verifier judgment, actor memory update,
        then curriculum review.
      \item If the preceding target passed and no new practice was
        selected, finish DRS. Otherwise, return to Step~3.
        A curriculum \texttt{STALLED} decision instead allows one final
        target attempt and learning update before recording the actual
        terminal verdict and ending DRS.
      \item \textbf{Evaluation:} freeze $M$, reset the environment,
        and execute $q$ with the actor--verifier harness and no learning.
        Archive the official score outside the agents and check that
        frozen host memory is unchanged.
    \end{enumerate}
    \noindent\textit{Guard:} unresolved verification or infrastructure
    errors suspend advancement; they are not converted into task
    verdicts. A completed lifecycle need not have a successful target
    verdict.\par\smallskip
    \hrule
  \end{minipage}
\end{center}


\section{Agent Prompt Templates}
\label{app:agent-prompts}

The prompts implement the role boundaries described in
Appendix~\ref{app:implementation} through explicit instructions for
execution, verification, exploration, and learning.
The boxes reproduce selected passages from the reference implementation,
not complete system messages; ellipses mark omitted passages.
Task instructions, current memory, project outcomes, and verifier
reports are supplied at runtime as described below.
The source prompts refer to BRS and DRS as Phase 1 and Phase 2,
respectively; their wording is retained in the excerpts.

\definecolor{RsiPromptActor}{HTML}{EFF5FC}
\definecolor{RsiPromptVerifier}{HTML}{EFF8F1}
\definecolor{RsiPromptBroad}{HTML}{FCF5E8}
\definecolor{RsiPromptDeep}{HTML}{ECF8F7}
\definecolor{RsiPromptDistill}{HTML}{F4F0FA}
\definecolor{RsiPromptReconcile}{HTML}{FCF0F1}
\definecolor{RsiPromptDiagnosis}{HTML}{F3F4F6}
\newtcolorbox{rsipromptbox}[2]{%
  enhanced,breakable,
  colback=#1,colbacktitle=#1,
  colframe=PrettyNavy!22,coltitle=PrettyText,coltext=PrettyText,
  boxrule=0.45pt,arc=2pt,
  left=8pt,right=8pt,top=6pt,bottom=6pt,
  before skip=9pt,after skip=10pt,
  fonttitle=\small\bfseries,fontupper=\small,
  title={#2}}

\subsection{Task Execution and Verification}
\label{app:execution-prompts}

\paragraph{Actor Instructions.}
The actor prompts combine program-based control with the task's literal
requirements. The shared system message defines the action interface,
while a practice charter supplies the project instruction and available
memory. Prompt P1 shows passages from these two components; it does not
include the model-dependent visual-action declarations or the task itself.

\begin{rsipromptbox}{RsiPromptActor}{Prompt P1. Program actions and task requirements}
You solve ONE task on a real Ubuntu machine by WRITING PROGRAMS. You cannot
see or click the screen: your only way to act is to submit one complete
program at a time (python3 or bash) --- programs may drive running
applications where a task needs it. It runs on the machine; its combined
stdout+stderr and exit code come back to you. Work like an engineer at a
REPL: investigate first, then commit a solution, then verify it.

[\ldots]

Code is your control channel, not a reinterpretation of the project: every
action you take is a submitted program, and those programs may inspect and
operate the machine or automate a required application. Choose the
implementation yourself. Whatever method you choose, the candidate must
satisfy the project's literal requirements, including any named-application,
native-editable-state, behavior, rendered-output, workflow, or provenance
requirement.
\end{rsipromptbox}

\paragraph{Verifier Instructions.}
The verifier prompt defines evidence-based acceptance independently of
the actor agent's self-assessment. Prompt P2 shows the target configuration
with unresolved-evidence reporting and checkpoint-protected inspection.
The authoritative task and candidate context are supplied separately;
practice configurations use the two-verdict interface described in
Appendix~\ref{app:agent-interfaces}.

\begin{rsipromptbox}{RsiPromptVerifier}{Prompt P2. Independent candidate verification}
Treat observations as evidence, not proof by assertion. Derive every binding
requirement from the authoritative task and falsify nearby plausible
substitutes. PASS only when every material requirement is affirmatively
supported. Publish FAIL when concrete evidence establishes a material
violation. Publish UNVERIFIED when a material claim remains unresolved after
investigation or credible instruments disagree.

[\ldots]

At the beginning of this inspection the harness saved a complete QEMU
checkpoint. You can observe its files, processes, localhost services,
network, GUI, and IPC and may run arbitrary tests. When this inspection ends,
the harness restores that checkpoint before the candidate is graded, so
none of your in-VM effects enter the scored state.
\end{rsipromptbox}

\subsection{Curriculum Generation}
\label{app:curriculum-prompts}

\paragraph{Broad Recursive Self-exploration.}
The BRS prompt directs the curriculum agent to acquire complementary
experience around the target query. Its runtime inputs include the exact
query, current memory, and completed wave outcomes. The resulting handoff
specifies a wave of independent projects or a stopping decision, with
project fixtures prepared separately.

\begin{rsipromptbox}{RsiPromptBroad}{Prompt P3. Broad Recursive Self-exploration}
The exact target query is disclosed only as a search direction. Derive
diverse prerequisite, variant, contrast, and stress projects around its
capability neighborhood. Do not reproduce or attempt the unchanged target
in Phase 1; its first exact attempt is reserved for Phase 2.

[\ldots]

Each project should improve reusable capability or discriminate an
important uncertainty; do not optimize for easy passes, episode count, or
memory volume. Projects in the same wave receive the same pre-wave memory
snapshot and cannot see or depend on sibling work or results. After the
complete wave returns, reconsider your hypotheses and freely choose the
next wave. When further target-relevant exploration has insufficient
expected value, you may end Phase 1. No fixed project count is a semantic
convergence rule.

[\ldots]

When you independently judge Phase 1 complete, publish decision "SATURATED"
with an empty projects list and your evidence in rationale. Use decision
"STALLED" only when no productive project can currently be authored, also
with an empty list.
\end{rsipromptbox}

\paragraph{Deep Recursive Self-exploration.}
The DRS prompt makes further practice conditional on what the preceding
experience reveals. In addition to the target query and memory, the
curriculum agent receives the grounded outcome and the actor agent's
learning diagnosis. A successful attempt may therefore lead to a
contrastive exercise, while a failed attempt can motivate practice that
distinguishes competing explanations.

\begin{rsipromptbox}{RsiPromptDeep}{Prompt P4. Deep Recursive Self-exploration}
After PASS, do not create practice by default: choose PROJECT only when a
contrast or stress case can test an important uncertain or overgeneralized
hypothesis. After FAIL, prefer practice that discriminates among plausible
capability gaps. You choose the next experience, not memory wording and not
target correctness.

[\ldots]

READY\_FOR\_TARGET means only that no additional learning experience
currently has greater expected value than returning control to the
unchanged target lifecycle.

[\ldots]

No action is tied to a fixed project or turn count.
\end{rsipromptbox}

\subsection{Memory Consolidation and Learning Handoffs}
\label{app:memory-prompts}

\paragraph{Experience Distillation.}
The distillation prompt asks the actor agent to learn from grounded
experience without treating a verdict as a complete causal explanation.
It is delivered in the context that performed the task, together with
the terminal outcome and full verifier report. The actor agent retains
control over memory contents and may also leave memory unchanged.

\begin{rsipromptbox}{RsiPromptDistill}{Prompt P5. Experience distillation}
Make causal claims only when your trajectory and evidence support them,
and preserve uncertainty where they do not.

[\ldots]

You may investigate remaining questions if the available project state
makes that useful, and you may add, revise, reorganize, delete, or leave
memory unchanged. You own the content, representation, retrieval strategy,
scope, and stopping decision; there is no required schema, length, number
of files, or number of turns. A FAIL may still contain valuable evidence,
but must not be recorded as a verified success.

Do not continue changing the terminal project for credit. Declare done
when the memory you choose to carry forward is ready.
\end{rsipromptbox}

\paragraph{Memory Reconciliation.}
The reconciliation prompt asks the actor agent to review the whole memory
bank before its update is promoted. It follows distillation in the same
context, so the task trajectory, verifier evidence, and draft memory remain
available. The instructions emphasize correcting existing advice as well
as adding new experience.

\begin{rsipromptbox}{RsiPromptReconcile}{Prompt P6. Memory reconciliation}
Your first-pass memory update is a draft and has not been promoted.

[\ldots]

Look for conflicting assertions, unsupported causal explanations, stale
environment assumptions, and conclusions broader than the observed evidence.
Investigate when that is useful; otherwise narrow or qualify claims,
preserve uncertainty, reorganize them, or remove them. Do not merely append
this episode while leaving contradicted older advice stated as fact.

[\ldots]

There is no required claim table, schema, report, length, number of files,
or number of turns.
\end{rsipromptbox}

\paragraph{Learning Diagnosis.}
The diagnosis prompt connects actor-owned learning to the next curriculum
decision. The actor agent writes a separate handoff containing the
conclusions and uncertainties it chooses to communicate, rather than
sharing its private reasoning transcript. This handoff informs experience
selection without prescribing a fixed curriculum.

\begin{rsipromptbox}{RsiPromptDiagnosis}{Prompt P7. Learning diagnosis for curriculum review}
Explain the hypotheses that now seem most useful for choosing the next
experience: attempted approaches, observed limitations, plausible causal
reasons, what appears reliable, what remains uncertain, and which
distinctions or stress cases could discriminate among competing explanations.

[\ldots]

Do not include private chain-of-thought or a turn-by-turn transcript:
provide only the concise conclusions and hypotheses you choose to
communicate. There is no required schema, length, or organization.
\end{rsipromptbox}


\section{Detailed Experimental Setup}
\label{app:experimental-setup}
\label{app:provenance}

\subsection{Benchmark Descriptions and Evaluation}
\label{app:benchmark-details}

\paragraph{OSWorld 2.0.}
OSWorld 2.0 evaluates long-horizon workflows in desktop applications
and task-facing web services~\cite{yuan2026osworld20benchmarkingcomputer}.
Tasks require agents to recover information from the environment,
coordinate dependent operations, and produce a correct final state or
artifact. Our experiments use the August 8, 2026 release and report
the 82-task offline subset rather than the full 108-task inventory.
This subset includes document editing, media production, and
specialized engineering and scientific software workflows.

\paragraph{Agents' Last Exam.}
ALE evaluates professional workflows that combine software interaction,
code execution, and the production of verifiable
deliverables~\cite{sun2026agentsexam}.
We use the 67-task Near-term inventory specified by the
\href{https://github.com/rdi-berkeley/agents-last-exam/blob/d10fb61a14f9719774c3520c5763068b28ef5546/selected_tasks/full/near-term.txt}
{pinned task manifest}, which includes computing, finance, engineering,
scientific analysis, and visual-media tasks.
Our aggregates cover all 67 Near-term tasks.

\begin{table}[htbp]
  \centering
  \caption{Benchmark coverage used for our reported aggregates.}
  \label{tab:benchmark-coverage}
  \small
  \setlength{\tabcolsep}{5pt}
  \renewcommand{\arraystretch}{1.12}
  \begin{tabularx}{\linewidth}{@{}lXrr@{}}
    \toprule
    Benchmark & Split & Inventory & Aggregate denominator \\
    \midrule
    OSWorld 2.0 & 0808 offline & 82 & 82 \\
    Agents' Last Exam & Near-term & 67 & 67 \\
    \bottomrule
  \end{tabularx}
\end{table}

\paragraph{Coverage Exceptions.}
Infrastructure-related missingness is recorded separately from task
performance. For OSWorld, T082 has no valid official score after a
setup failure; it is counted as zero in both reported aggregates by
the chosen reporting convention. The resulting coverage is summarized
in Table~\ref{tab:benchmark-coverage}.

\paragraph{Metrics.}
Partial and Binary measure complementary aspects of task completion.
For normalized task scores $s_i\in[0,1]$ over $N$ included tasks, we
compute $\mathrm{Partial}=100\sum_{i=1}^{N}s_i/N$ and
$\mathrm{Binary}=100\sum_{i=1}^{N}\mathbf{1}\{s_i=1\}/N$.
Each task has equal weight, and the baseline and RSI aggregates use
the same denominator within each benchmark. Partial captures graded
progress, whereas Binary requires full credit on a task.

\subsection{Agent Harness and Configuration}
\label{app:harness-configuration}

\paragraph{Shared Execution Harness.}
The baseline and frozen-memory RSI evaluation use the same
code-as-policy actor--verifier harness.
The actor agent executes programs and revises its candidate using
the verifier agent's findings. The verifier agent maintains a separate
context and returns \texttt{PASS}, \texttt{FAIL}, or
\texttt{UNVERIFIED}; an unverified outcome requests additional evidence
instead of being treated as success. RSI adds curriculum-guided
exploration and reusable memory, while the baseline disables
exploration and persistent memory. Model weights remain fixed.

\paragraph{Reference Configuration.}
Table~\ref{tab:agent-configuration} records the checked OSWorld
reference configuration and its two exploration stages.
GLM-5.3 is the default actor agent, and Kimi-K3 supplies the verifier
agent, curriculum agent, and visual observations.
The verifier and curriculum agents use separate contexts despite
sharing a model. The target query guides the curriculum agent in both
stages: Broad Recursive Self-exploration (BRS) acquires related
experience before Deep Recursive Self-exploration (DRS) attempts
the target itself.

\begin{table}[htbp]
  \centering
  \caption{Reference agent configuration and exploration settings.}
  \label{tab:agent-configuration}
  \label{tab:provenance}
  \small
  \setlength{\tabcolsep}{5pt}
  \renewcommand{\arraystretch}{1.12}
  \begin{tabularx}{\linewidth}{@{}
    >{\raggedright\arraybackslash}p{0.30\linewidth}
    >{\raggedright\arraybackslash}X@{}}
    \toprule
    Component & Setting \\
    \midrule
    Actor agent & GLM-5.3 by default \\
    Verifier agent & Kimi-K3; separate persistent context \\
    Curriculum agent & Kimi-K3; separate context with the target query as reference \\
    BRS & Nominal budget of eight projects; up to four execute concurrently \\
    Broad-stage budget check & After completed waves; ongoing waves are not truncated \\
    DRS & Sequential, with \texttt{curriculum\_review} stopping \\
    Primary-call sampling & Temperature $1.0$; top-$p$ $1.0$ \\
    Response limit & 65,536 tokens per primary model call \\
    Baseline memory & Disabled \\
    RSI evaluation memory & Frozen after exploration; no writeback \\
    Model parameters & Fixed throughout exploration and evaluation \\
    \bottomrule
  \end{tabularx}
\end{table}

\paragraph{Exploration and Stopping.}
The two stages use different stopping mechanisms to match their
exploration structure. BRS checks its nominal eight-project
budget only after a complete wave, so the realized project count can
exceed eight; four is the concurrency limit, not a required wave width.
DRS uses \texttt{curriculum\_review}: verified outcomes
inform memory updates and the curriculum agent's next decision.
A successful practice or target attempt does not by itself terminate
this process. If additional practice is selected, another target
attempt follows; exploration ends when the curriculum agent determines
that no further useful practice is needed. Earlier
\texttt{verifier\_pass} runs use a distinct stopping-policy variant.

\paragraph{Execution Safeguards.}
Numerical execution limits bound individual agent runs rather than
define a fixed number of RSI rounds. The OSWorld reference target actor
uses a 500-iteration limit and a 36,000-second watchdog per run.
Practice and curriculum configurations use 2,000 iterations and
86,400 seconds; target-script execution is limited to 600 seconds
per call. Consequently, the ten-hour target watchdog is not a
ten-hour ceiling on the complete exploration lineage.
The target harness also permits one stall-triggered role switch,
using Kimi-K3 as the actor agent and GLM-5.3 as the verifier agent
when execution budget remains.

\paragraph{Verification and Evaluation.}
Local verification is separated from the final benchmark evaluator.
In the OSWorld reference harness, verifier probes operate on a
checkpoint-protected copy of the candidate environment, which is
restored after inspection. After exploration, the environment is reset
and accumulated memory is reused without writeback for target
evaluation. The official evaluator is invoked after the actor--verifier
loop and does not supply scores or hidden checks to the learning agents.
The reference evaluator uses GPT-5.4 where model-based grading is
required, and the configured user simulator uses GPT-4o.


\subsection{RSI Task Selection and Reporting}
\label{app:rsi-task-selection}

\paragraph{Selection Rationale.}
We direct additional exploration toward tasks with remaining room for
improvement. In the documented OSWorld expansion cohort, tasks were
selected when their recorded baseline score was below full credit and
they were not already covered by an RSI lineage. Baseline full-score
tasks were not assigned additional exploration in this cohort, but
remain in the benchmark aggregates. A missing or invalid baseline was
not treated as evidence of an unsolved task for selection.

\paragraph{Documented Selection Cohort.}
The expansion cohort provides an explicit record of how the selection
rule was applied before new runs began.
Table~\ref{tab:rsi-selection-cohort} lists its task groups from the
September 7, 2026 selection snapshot. It distinguishes selected tasks from
previously launched lineages and from tasks receiving no new exploration.
These groups describe this cohort only, not the complete set of
OSWorld and ALE results included in Table~\ref{tab:main-results}.

\begin{table}[htbp]
  \centering
  \caption{Task selection in the documented OSWorld expansion cohort.}
  \label{tab:rsi-selection-cohort}
  \small
  \setlength{\tabcolsep}{5pt}
  \renewcommand{\arraystretch}{1.12}
  \begin{tabularx}{\linewidth}{@{}
    >{\raggedright\arraybackslash}p{0.27\linewidth}
    r >{\raggedright\arraybackslash}X@{}}
    \toprule
    Selection status & Tasks & Task IDs \\
    \midrule
    New exploration & 17 & T076, T077, T079, T081, T085, T087,
      T089, T090, T091, T092, T093, T098, T100, T101, T104, T105, T107 \\
    Existing RSI lineages & 5 & T080, T102, T103, T106, T108 \\
    Baseline full credit & 4 & T073, T084, T088, T094 \\
    Invalid baseline & 1 & T082 \\
    \bottomrule
  \end{tabularx}
\end{table}

\paragraph{Selection-Time Scores.}
Selection records and subsequently corrected scores serve different
purposes. T100 entered this expansion with a recorded baseline of zero;
the current main table uses its later corrected baseline and RSI values
of one. T082 was excluded from this expansion because it lacked a valid
baseline, although the benchmark aggregate counts it as zero.
Neither convention changes the original selection record.

\paragraph{Aggregation.}
The reported RSI aggregate retains both improvements and regressions.
For each task with a reported non-diagnostic RSI entry, that score
replaces the baseline; otherwise, the baseline score is retained.
Tasks that receive no additional exploration therefore remain in the
benchmark denominator, and retained baseline scores are not independent
RSI evaluations. The counts of scored RSI entries below describe the
main-table reporting set, not the number of eligible or launched tasks.
Recorded retries and selected checkpoints retain their reporting
qualifications in Appendix~\ref{app:main-table-reporting}.


\subsection{Main-Table Reporting Details}
\label{app:main-table-reporting}

\paragraph{Metrics and Sources.}
Table~\ref{tab:main-results} reports mean partial credit (Partial)
and full-task success rate (Binary), both as percentages.
A dash indicates an unavailable or unverified result.
Comparison scores come from the
\href{https://osworld-v2.xlang.ai/}{OSWorld} and
\href{https://agents-last-exam.org/leaderboard}{ALE} leaderboards,
except GPT-6 Astra's OSWorld result, which is reported by
\href{https://openai.com/index/gpt-6-astra/}{OpenAI}.
Partial and Binary are paired within the same published configuration.
For ALE, each model uses its highest-partial-score configuration
covering all 67 tasks.
Sources were accessed on September 11, 2026.
Published systems retain their original harnesses and execution budgets;
the cross-system comparison does not use a matched evaluation protocol.

\paragraph{OSWorld 2.0.}
Our provisional OSWorld aggregates cover 82 offline tasks,
with T082's setup failure counted as zero.
The RSI row replaces baseline scores with 41 reported
non-diagnostic RSI results, including recorded retry scores
and regressions; the remaining tasks retain their baseline scores.
Selected-run budgets and evaluation scopes are not fully matched.

\paragraph{Agents' Last Exam.}
Our ALE aggregates cover all 67 Near-term tasks.
The RSI row combines 19 reported RSI-column scores and
48 baseline scores.
These records include local corrected grades, ECG results
qualified by public-label transfer, and a separate no-BRS
Tax Form variant.
The aggregate is not a matched-protocol comparison with
published leaderboard results.
On Binary, RSIAgent scores 50.75, compared with 52.24 for GPT-6 Astra;
the partial-credit advantage does not extend to every reported metric.

\subsection{Four-Task Stage-Ablation Protocol}
\label{app:ablation-protocol}

\paragraph{Task selection and aggregation.}
The stage comparison uses four tasks from an exploratory
cohort selected for improvements over recorded baselines: T080
(WPS spreadsheet repair), T085 (REAPER radio-bumper editing), T089
(browser-based presentation repair), and T106 (3D Slicer liver
segmentation). The source-verified executions use GLM-5.3 for the actor
agent. The w/o RSI condition uses the recorded historical baseline,
and the single-stage conditions use their recorded scores.
Full RSI is the mean of two historical frozen-memory evaluations:
0.6869 and 0.6869 for T080, 0.9417 and 0.9413 for T085,
0.7950 and 0.8300 for T089, and 0.4976 and 0.5836 for T106.
Scores in this appendix use the 0--1 scale; Figure~\ref{fig:ablations}
expresses the same scores as percentages.

\paragraph{Single-stage conditions and stopping.}
The w/o DRS condition evaluates an exact copy of the memory acquired
through Broad Recursive Self-exploration (BRS), without Deep Recursive
Self-exploration (DRS). The w/o BRS condition starts DRS with empty
memory and permits at most two curriculum
practice projects; target attempts and their memory updates are not
counted as practice projects. T080 and T085 stop at the two-project
cap, while T089 reaches the curriculum-ready state after one project.
The archived T106 w/o BRS checkpoint also contains two practice projects.
The checkpoint records give the actor agent access to persistent
memory but do not give the curriculum agent direct memory access.
T085 uses the completed infrastructure-replacement w/o BRS run while
retaining its original w/o DRS evaluation. The resulting memory is
frozen before evaluation; each source-verified single-stage result records
one successful official grading call, no failed or corrected grade, and
no official feedback to the agent or memory writeback.

\section{Autonomous Game Development Setup}
\label{app:game}

We instantiate RSIAgent in the autonomous game-development setting using GameCraft-Bench~\citep{luo2026gamecraft}, which evaluates complete playable games through executable interaction traces and rubric-based judging. This setting is also related to continual game generation studied by Play2Code~\citep{huang2026gui}, where game development proceeds through repeated playtesting and code revision.

We randomly sample 40 tasks and follow the GameCraft-Bench evaluation pipeline. For each game, GLM-5.3-Flash serves as the development actor and verifier, Qwen3.8-27B serves as the playtesting agent, and Qwen3.8-27B is used for final evaluation. Each development run contains at most 15 iterative improvement rounds, with a maximum budget of 26 tool calls for the actor in each round. We compare Play2Code, RSIAgent without recursive experience evolution, and the full RSIAgent.

\section{Case Studies: From Memory Growth to Better Task Execution}
\label{app:memory-case-studies}

This appendix follows three OSWorld 2.0 trajectories from exploratory
practice to memory reuse. We first examine how exploration branches turn
observations into persistent records and reusable procedures, and then
compare execution without memory against execution with the accumulated
memory. T065 (railway booking) makes the acquisition process visible;
T049 (presentation repair) connects a recorded failure, subsequent
practice, and a corrected output; T044 (video editing) shows how an
application-specific procedure changes both the saved project and its
score. The examples include screenshots and artifact views retained
from the original runs.

A complementary FreeCAD case in Appendix~\ref{app:case-cad} traces how
drawing interpretation and geometric checks are retained and reused
when reconstructing a mechanical support bracket.
Appendix~\ref{app:case-reaper} adds a REAPER audio-production case,
connecting fragment interpretation and rendering practice to memory
reuse and higher partial credit on T085.

Following the main text, we organize the cases around
\emph{Broad Recursive Self-exploration} (BRS),
\emph{Deep Recursive Self-exploration} (DRS), and
\emph{Test-time Memory Reuse}. BRS builds broad environment knowledge;
DRS refines it through feedback-guided practice; and test-time memory
reuse applies the frozen corpus to task execution
(Section~\ref{sec:evaluation}). Model weights remain fixed throughout.
In these runs, DRS includes practice on the target itself. The final
comparisons pair RSIAgent with its historical memory-free baseline,
RSIAgent (w/o RSI).

\subsection{Memory Accumulates Through Exploration and Consolidation}
\label{app:case-memory-growth}

\paragraph{From an empty directory to a working knowledge base.}
All three trajectories begin with empty memory. During BRS, the
exploration projects cover different aspects of the surrounding task
family. Each completed project contributes an outcome and a memory
update. Some updates introduce a new topic or episode record; others
extend an existing procedure, correct a previous interpretation, or
refresh the index that a future actor will use to find relevant material.
The resulting memory contains both concrete precedents and instructions
for handling the next task.

\begin{figure}[!htbp]
  \centering
  \includegraphics[width=\linewidth]{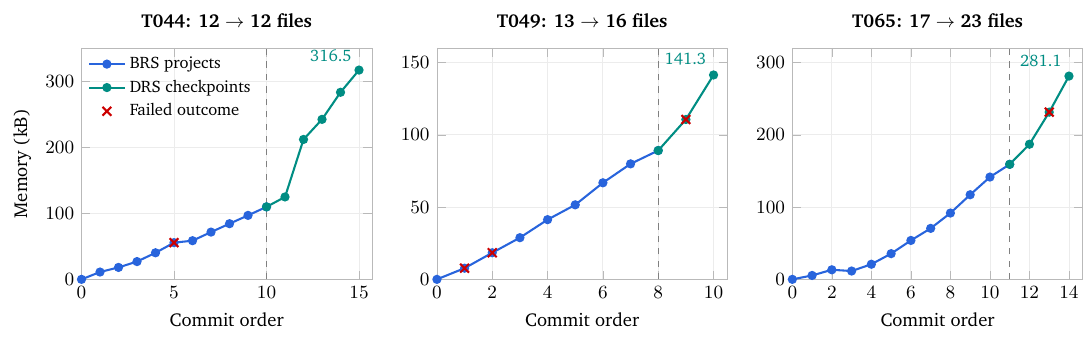}
  \caption{Memory accumulation in the three case studies. Blue points
  count committed BRS projects; green points count DRS target-cycle
  checkpoints, including any intervening practice. The dashed line marks
  the transition between these two counting units. Red crosses mark
  failed outcomes that still produced memory updates. File counts above
  each panel compare the end of BRS with the final freeze. Size is
  measured in decimal kB. This plot is reconstructed from the saved
  journals; the subsequent figures show retained runtime images.}
  \label{fig:case-memory-growth}
\end{figure}

Figure~\ref{fig:case-memory-growth} follows the saved memory checkpoints.
T044 completes ten BRS projects and accumulates 109,647 bytes in
12 files; T049 completes eight projects and accumulates 89,086 bytes in
13 files; and T065 completes eleven projects and accumulates 159,065 bytes
in 17 files. DRS builds on these same corpora. At the final freeze,
they contain 316,541, 141,312, and 281,085 bytes, respectively. The archive
contains 39 journal-linked snapshots across the three trajectories, with
file counts and byte totals checked against the journals. The final
evaluation records match the frozen DRS memory in file count, byte
count, and tree hash.

\begin{table}[!htbp]
  \centering
  \caption{Memory checkpoints and practice volume. Sizes are the summed
  bytes of regular files in each memory snapshot. A target cycle and a
  supplementary project are separate units of practice.}
  \label{tab:case-memory-checkpoints}
  \small
  \begin{tabularx}{\linewidth}{@{}lrrrX@{}}
    \toprule
    Task & BRS projects & BRS bytes & Frozen bytes & DRS practice \\
    \midrule
    T044 & 10 & 109,647 & 316,541 & 5 target outcomes; 7 supplementary projects \\
    T049 & 8 & 89,086 & 141,312 & 2 target outcomes; 1 supplementary project \\
    T065 & 11 & 159,065 & 281,085 & 3 target outcomes; 2 supplementary projects \\
    \bottomrule
  \end{tabularx}
\end{table}

Growth also takes the form of revision within a stable set of files.
For example, T044 retains twelve files throughout DRS while its
memory expands from about 110 to 317 kB: the actor keeps enriching the
same editing manuals, verification procedures, and failure records.
Conversely, T065's third BRS consolidation reduces memory from
13,243 to 11,510 bytes. The history therefore captures both accumulation
and reorganization rather than simple concatenation of every interaction.

\paragraph{Different branches contribute complementary lessons.}
The T049 BRS index gives a concrete view of what is being accumulated.
One PDF-audit project fails because the actor applies a familiar
coordinate-system convention instead of the conversion explicitly given
in the task. A later audit in the same family uses the stated mapping
and obtains an exact coordinate match. Presentation projects explore
detached connectors, overflow repair, scaled reference figures, and
alignment against intact elements. For example, the callout project
derives connector anchors from a PDF through a scale of 8,000 EMU per
point and validates the mapping against a declared-correct connector
before repairing the broken ones. These experiences populate separate
coordinate-mapping notes, a PPTX editing playbook, interpretation rules,
and project records.

The index itself also learns from the branch structure. One episode
finds that sibling records have appeared since its initial inspection;
rewriting the index from its earlier view would leave those records
unlisted. Its retained lesson is to enumerate the current project-record
directory and topic headings again at phase boundaries before updating
the index. Thus, the accumulated knowledge includes how to preserve and
retrieve discoveries made by other branches, in addition to how to
operate the application.

\paragraph{A failed export becomes a reusable check.}
T044's fifth BRS project asks for an untouched Shotcut timeline and a
default H.264/MP4 export. The project receives an internal FAIL after the
verifier compares the exported audio settings with a command-line render
and reports a default-setting discrepancy. The subsequent consolidation
creates \texttt{shotcut-default-export-fail.md} and
\texttt{media-verification-checklist.md}, modifies nine existing files,
and increases memory from 40,186 to 55,506 bytes. The retained editing
manual distinguishes the GUI export settings from the bundled engine's
no-override defaults and calls for checking the actual consumer settings
and output streams. This episode illustrates how a problematic outcome
can still enrich the procedures used in later work: a plausible-looking
video is accompanied by explicit checks of its export configuration.

\begin{figure}[!htbp]
  \centering
  \begin{minipage}[t]{0.49\linewidth}
    \centering
    \includegraphics[width=\linewidth]{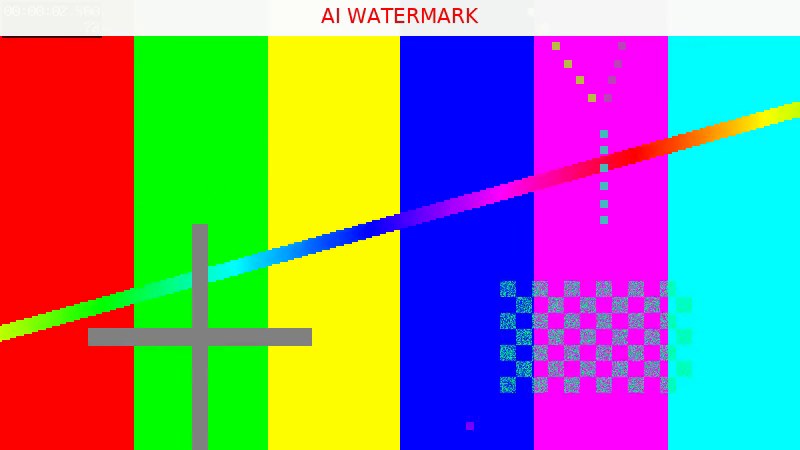}
    \par\smallskip\small (a) Project 1 input: $800\times450$.
  \end{minipage}\hfill
  \begin{minipage}[t]{0.49\linewidth}
    \centering
    \includegraphics[width=\linewidth]{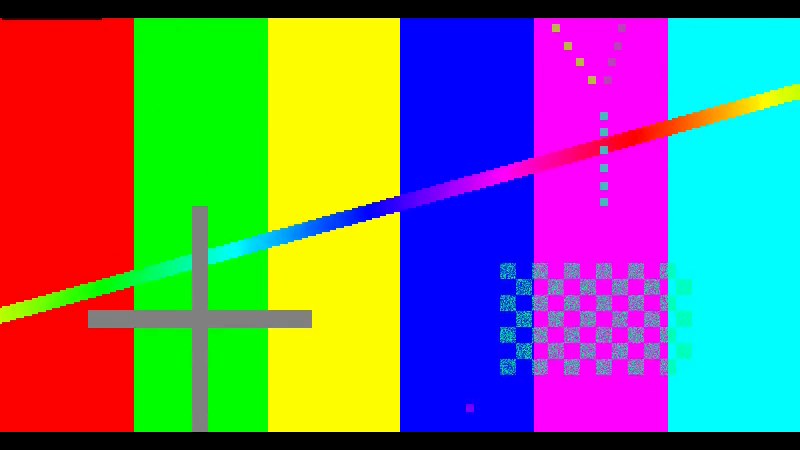}
    \par\smallskip\small (b) Project 1 output: 18-pixel bars.
  \end{minipage}
  \par\medskip
  \begin{minipage}[t]{0.49\linewidth}
    \centering
    \includegraphics[width=\linewidth]{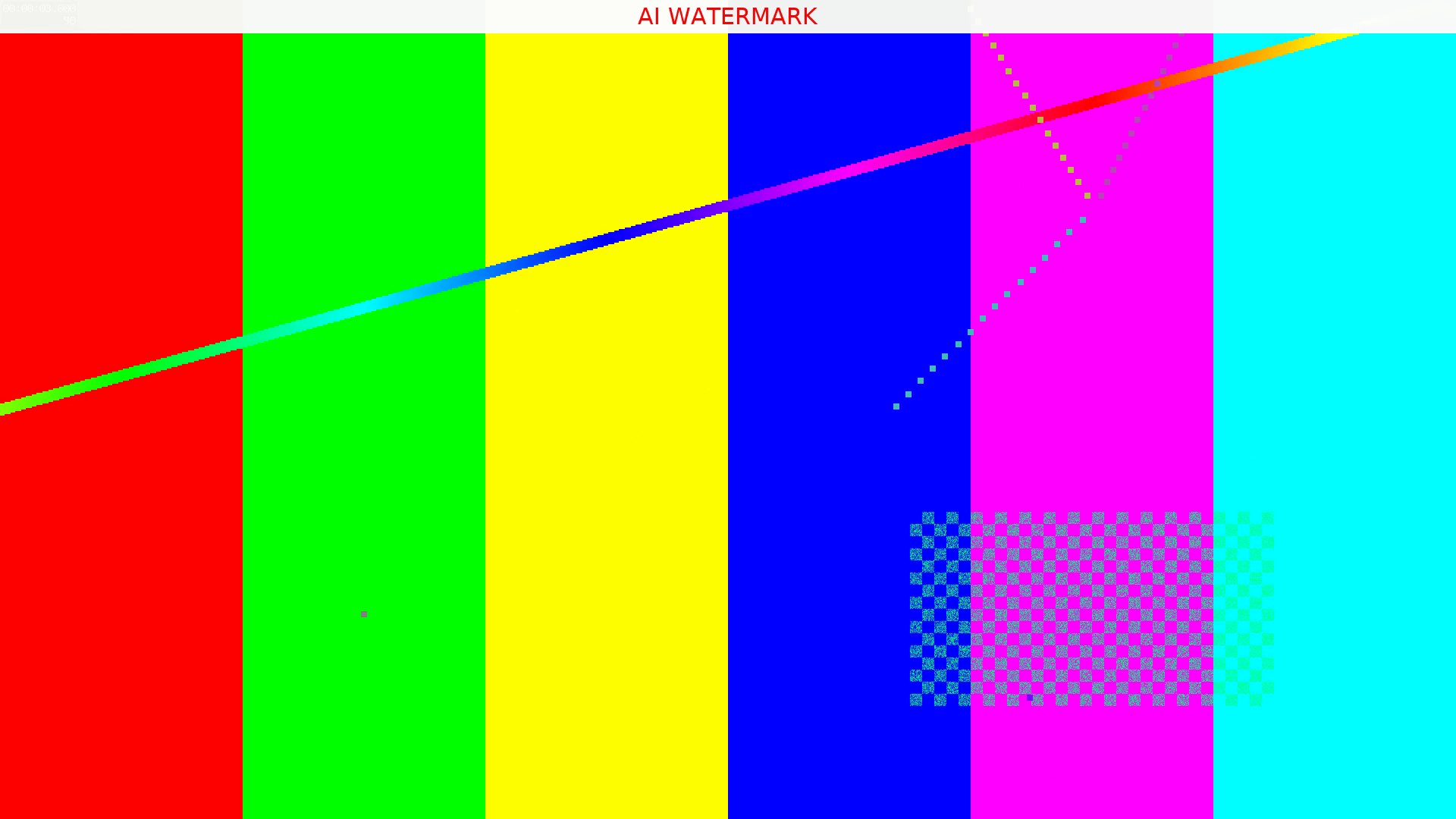}
    \par\smallskip\small (c) Project 7 input: $1920\times1080$.
  \end{minipage}\hfill
  \begin{minipage}[t]{0.49\linewidth}
    \centering
    \includegraphics[width=\linewidth]{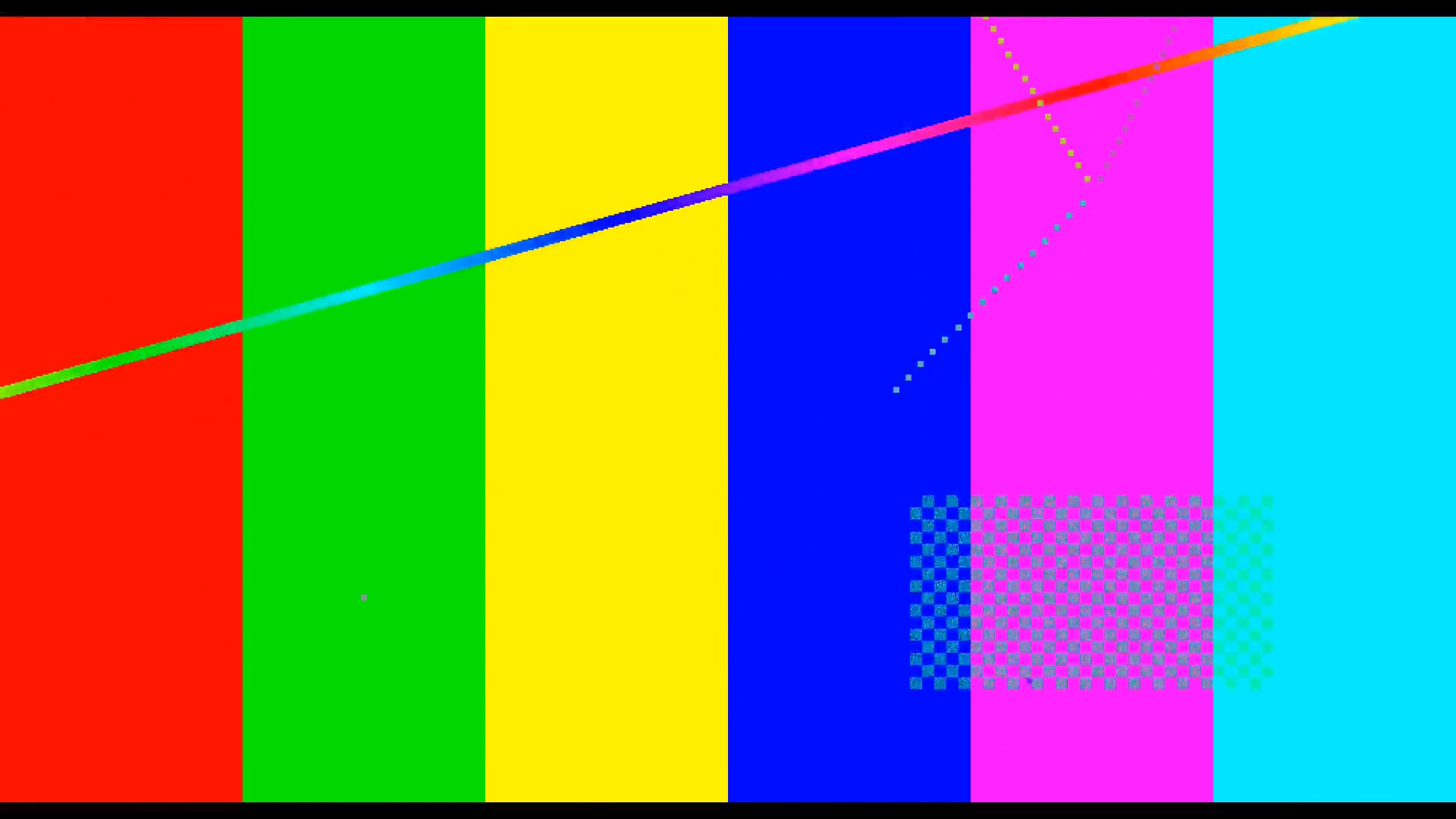}
    \par\smallskip\small (d) Project 7 output: 22-pixel bars.
  \end{minipage}
  \caption{Learning a reusable video-editing procedure through BRS.
  Each row shows input and output frames inspected in the same T044
  practice project. The top watermark band disappears, while the color
  bars and geometric patterns provide landmarks for checking image
  proportions. The two projects use different resolutions and crop
  heights, supplying repeated experience for the editing manual.}
  \label{fig:case-brs-video-pairs}
\end{figure}

\paragraph{Repeated visual checks across video resolutions.}
Other T044 branches establish the crop-and-letterbox procedure itself.
Figure~\ref{fig:case-brs-video-pairs} pairs the input and output frames
from an $800\times450$ exercise and a later $1920\times1080$ exercise.
The measured top bands are 36 and 44 pixels, respectively, and the
corresponding outputs have symmetric 18- and 22-pixel bars. The verifier
also checks the image mapping and output streams, so the procedure is
grounded in both the visible result and the saved video. The later
episode is recorded in the editing manual as another confirmation that
a top-only crop can remain centered and unscaled when the profile width
is preserved. By this point, the memory describes a method tested on
several clips, with the crop height measured anew for each input.

\FloatBarrier
\subsection{A Branch in Detail: Learning to Observe Changing Seat Availability}
\label{app:case-seatwatch}

T065's eighth BRS project isolates one prerequisite of railway
booking: observing a changing seat map and choosing the first release
that satisfies the requested seating rule. The actor opens a local
practice page in Chrome and must watch the rendered interface for at
least 100 seconds. In this exercise, the two travelers need seats in the
same row, one D and one F, simultaneously available. The actor must
record the releases, their first observed times, and the earliest valid
choice. Page information must come from the GUI.

\begin{figure}[!htbp]
  \centering
  \begin{minipage}[t]{0.48\linewidth}
    \centering
    \includegraphics[height=3.55in,trim=70bp 170bp 1450bp 25bp,clip]{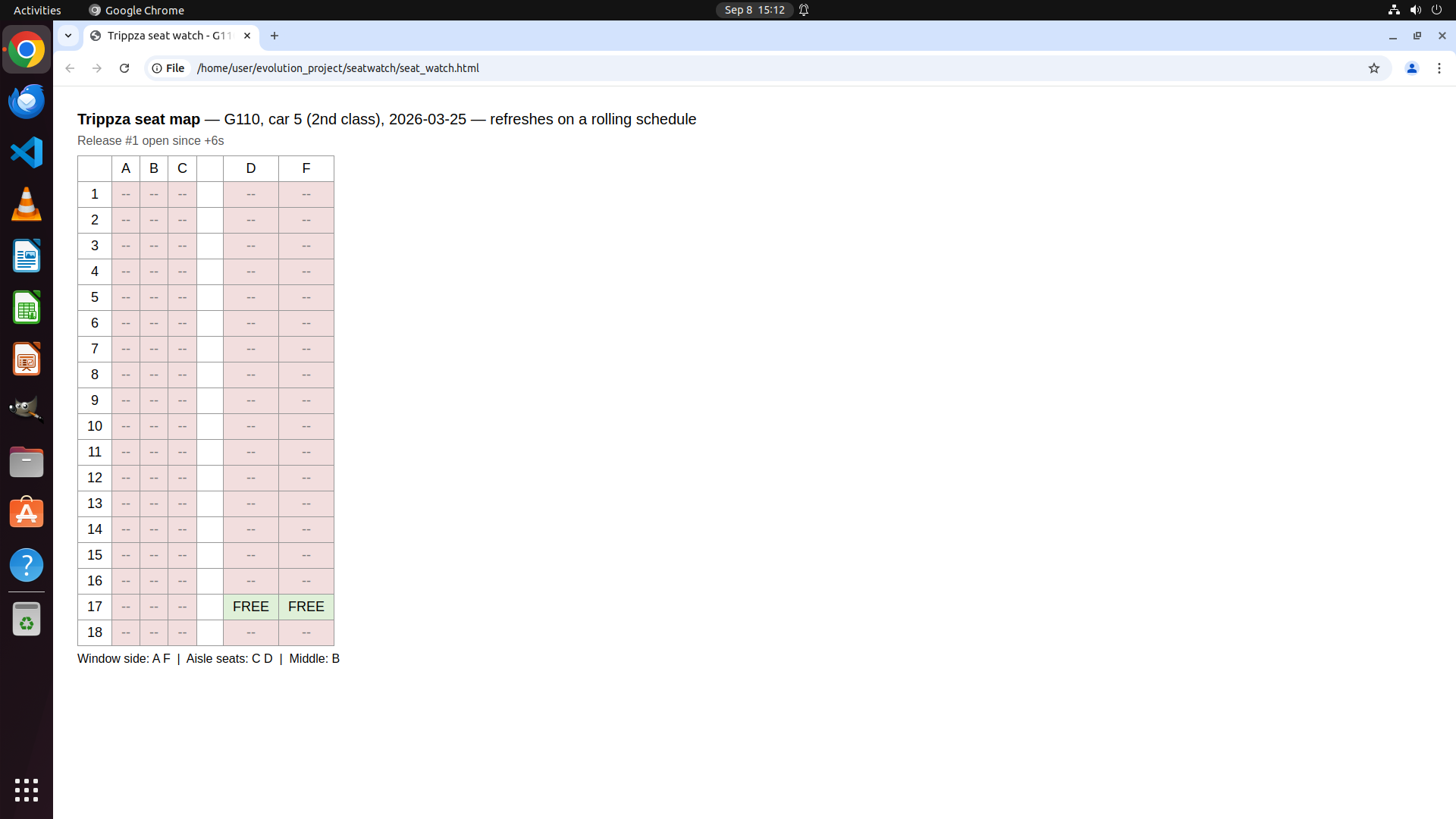}
    \par\smallskip\small (a) Release 1: 17D and 17F are free.
  \end{minipage}\hfill
  \begin{minipage}[t]{0.48\linewidth}
    \centering
    \includegraphics[height=3.55in,trim=70bp 170bp 1450bp 25bp,clip]{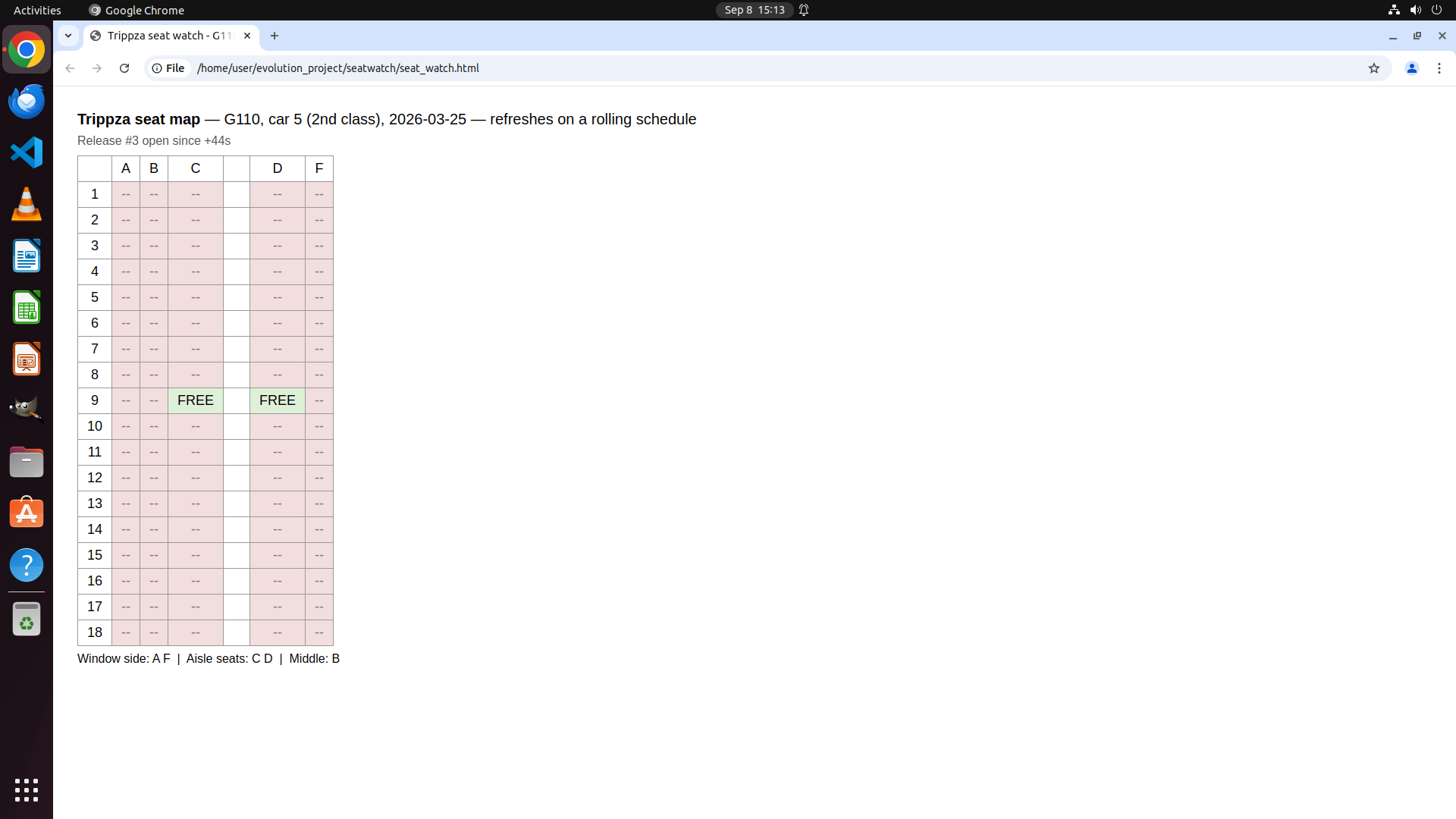}
    \par\smallskip\small (b) Release 3: 9C and 9D are free.
  \end{minipage}
  \caption{Learning from a dynamic interface in T065, BRS, episode 8.
  These are crops of the original Chrome screenshots of the local
  practice page, retaining the browser chrome and seat table. The first
  release satisfies this exercise's D/F rule; the third does not. Notice
  that the wider columns move with the locations of the \texttt{FREE}
  labels. The original full screenshots are retained with the figure
  assets.}
  \label{fig:case-seatwatch}
\end{figure}

\paragraph{What the branch observes.}
The page exposes five successive release windows. The first makes
17D and 17F available, so it is already a valid choice. The second
offers only 8A; the third offers 9C and 9D; the fourth offers only 14F;
and the fifth offers 11D, 11F, and 12A. Although the fifth window also
contains a valid D/F pair, it arrives later. The third window is a useful
contrast: two seats in the same row do not satisfy the exercise when one
of their letters is outside the allowed set.
Figure~\ref{fig:case-seatwatch} shows the first and third releases as
captured during the run.

\paragraph{What goes wrong, and what is learned.}
The challenge extends beyond reading the seat letters. The table sizes
its columns from their contents: a column containing \texttt{FREE} is
about 72 pixels wide, whereas an occupied-only column is about 36 pixels
wide. Consequently, seat positions shift as the release changes. A
fixed pixel grid initially misclassifies the third and fourth releases.
The actor corrects the readings by detecting the columns separately in
each state, using the wider aisle gap between C and D as a structural
cue. The two screenshots make the reason for this correction visible:
the widened columns are D/F in the first state and C/D in the third.

The branch also encounters a monitoring failure. Its detached watcher
fails twice at compilation, leaving the early releases unobserved. The
actor restarts the deterministic practice page in a fresh browser
profile and switches to an inline capture loop. The completed record
contains 663 frames over approximately 335 seconds, together with
timestamps and the first observed frame for each release. The verifier
checks the recorded seat sets against the page's release schedule and
independently inspects the captured frames, returning PASS.

\paragraph{How the observation becomes memory.}
Consolidation creates \texttt{seatwatch\_tasks.md} and an episode record,
and extends the existing observation and environment playbooks. The
memory preserves the specific release sequence as well as the reusable
method: distinguish simultaneous availability from seats seen at
different times, test the full seating condition, re-detect layout after
state changes, and confirm that the monitoring process is actually
producing evidence. The episode record states the key discovery directly:
\begin{quote}
  \small
  ``content-driven table column widths (FREE \textasciitilde72 px,
  occupied \textasciitilde36 px)'' and ``a frozen grid misclassified
  releases \#3/\#4 until columns were re-detected per state''.
\end{quote}
This commit grows the corpus from ten files and 70,513 bytes to twelve
files and 91,697 bytes. The additional memory is therefore tied to an
observed failure, a corrected method, and a verified decision.

\begin{figure}[p]
  \centering
  \includegraphics[width=0.96\linewidth,trim=70bp 380bp 460bp 25bp,clip]{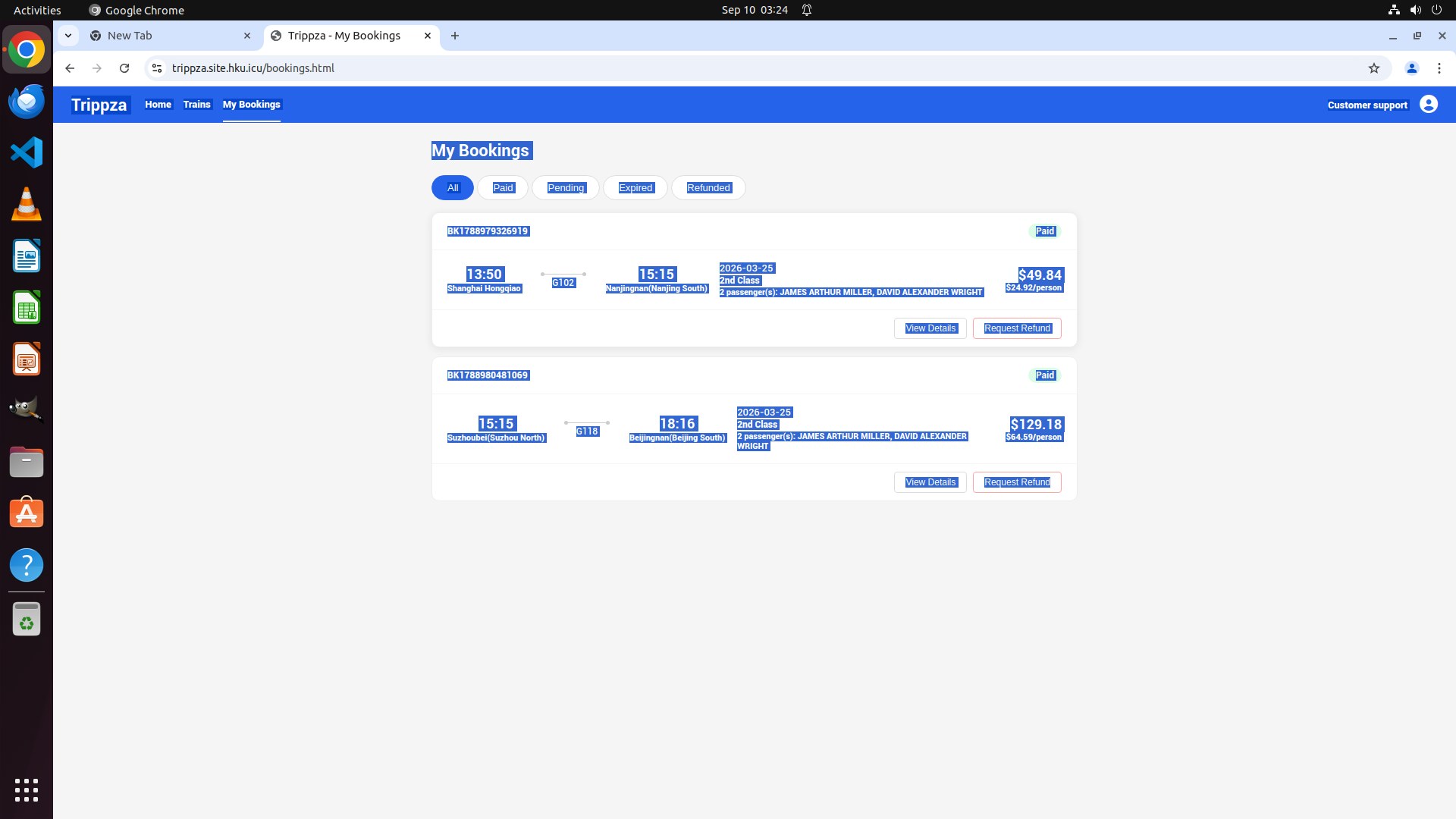}
  \par\smallskip\small (a) DRS cycle 2: G102 and G118; internal FAIL on transfer interpretation.
  \par\medskip
  \includegraphics[width=0.96\linewidth,trim=70bp 380bp 460bp 25bp,clip]{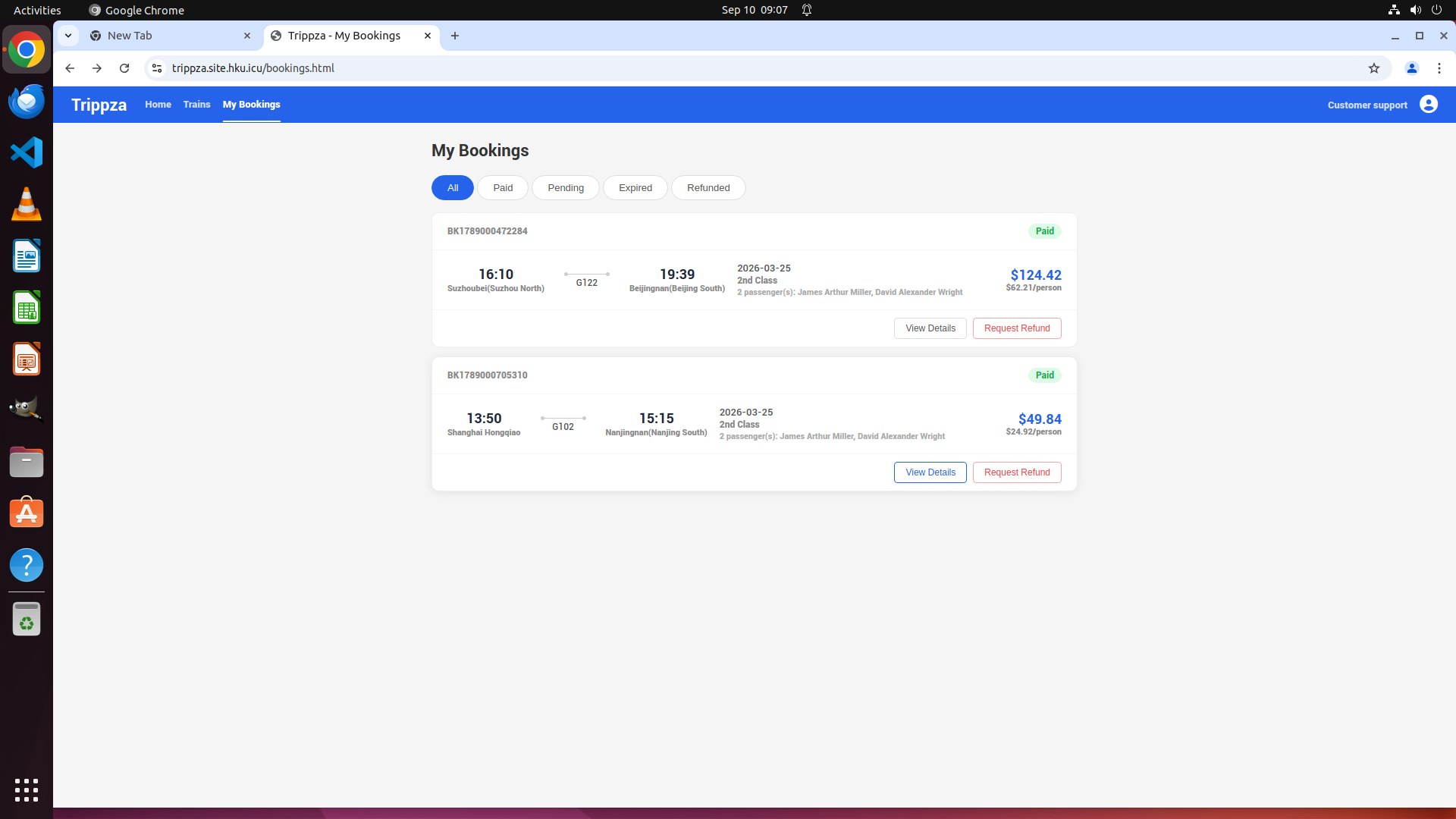}
  \par\smallskip\small (b) DRS cycle 3: G102 and G122; intermediate boarding accepted.
  \caption{T065's booking-state evidence during DRS. The original
  Chrome screenshots are cropped to retain the browser controls and
  both Paid orders. In cycle 2, the verifier evaluates the purchased
  endpoints as disconnected. In cycle 3, the verifier accepts boarding
  G122 at its intermediate Nanjing South stop, as documented in the
  accompanying diagnosis. These are successive practice cycles; the
  final test-time itinerary is described separately below.}
  \label{fig:case-drs-bookings}
\end{figure}

\paragraph{DRS extends observation rules into booking decisions.}
The later target cycles expose another distinction: a completed payment
and a valid itinerary require different checks.
Figure~\ref{fig:case-drs-bookings} shows the actual My Bookings pages
from cycles 2 and 3. Cycle 2 contains Paid orders for G102 and G118,
but receives an internal FAIL when the verifier requires the purchased
segments to connect directly. This outcome enters memory as a stricter
transfer rule. In cycle 3, the actor revisits that rule against the task's
explicit permission to buy a longer ticket and board at an intermediate
stop. It books G102 and G122; this verifier accepts boarding the second
train at Nanjing South, with a 1-hour-39-minute transfer and a total
price of \$174.26. Memory records the successful variant alongside the
earlier failure, refining the conditions under which a route should be
chosen. The two verdicts reflect different interpretations of the
transfer requirement during DRS.

The same cycle discovers a further observation failure: a scripted
clipboard read can return incomplete results while the page is still
rendering. The screen shows train cards, but the text read contains
none. The resulting memory adds a completeness check before treating
an empty read as unavailable inventory. This extends the earlier
seat-table lesson from locating changing columns to confirming that a
dynamic page has finished presenting the state being measured.

\begin{figure}[!htbp]
  \centering
  \includegraphics[width=0.95\linewidth,trim=360bp 510bp 360bp 45bp,clip]{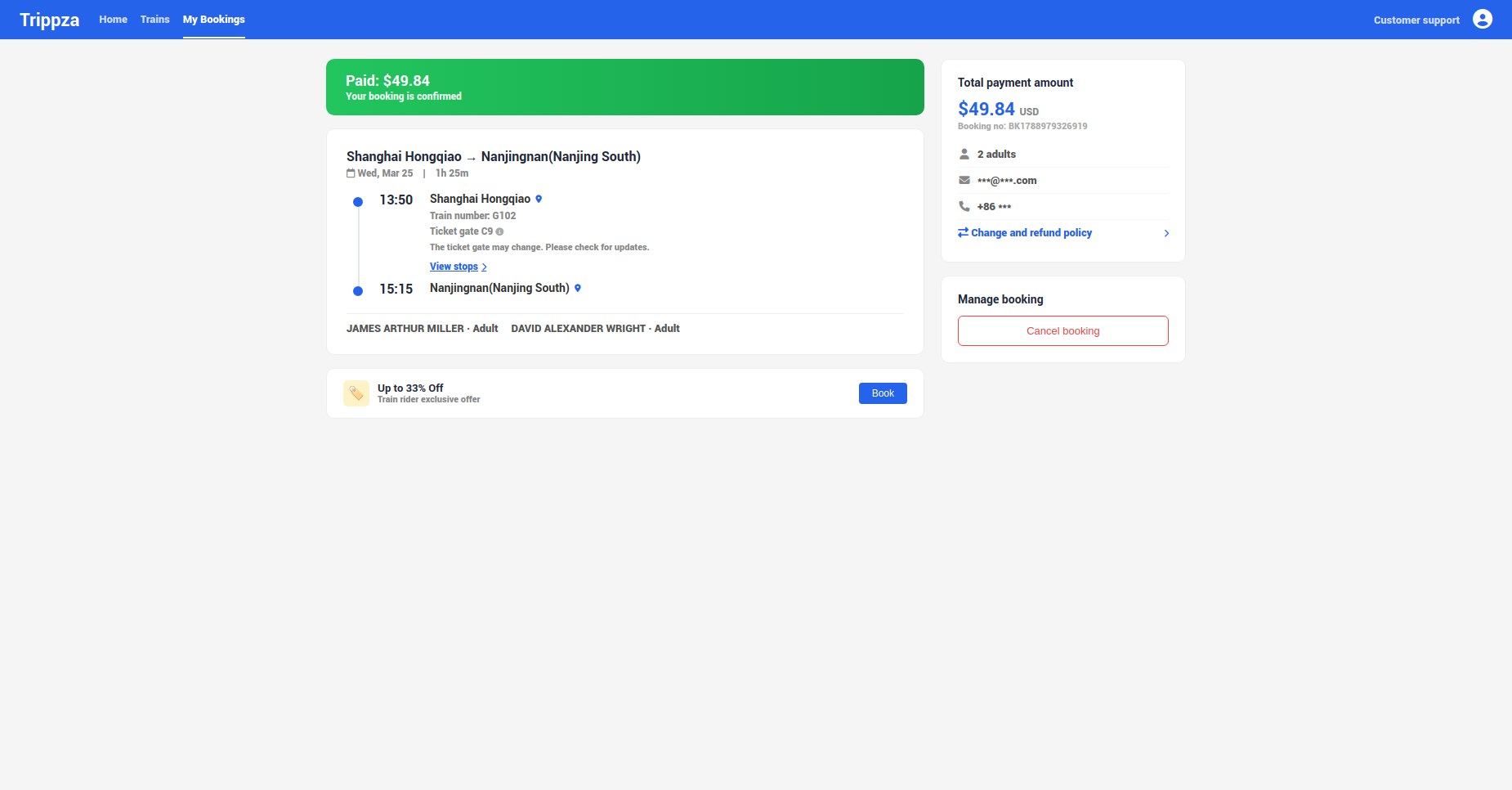}
  \caption{T065, DRS cycle 2: the saved G102 booking detail inspected
  during the run. The Paid banner, train times, two passengers, and
  payment summary provide persistent evidence of this completed order.
  The screenshot is cropped around the booking panels.}
  \label{fig:case-drs-booking-detail}
\end{figure}

\paragraph{Checking the persistent booking state.}
The detail page supplies a closer view of what the interface confirms
(Figure~\ref{fig:case-drs-booking-detail}): G102, the two passengers,
the Shanghai--Nanjing segment, and the paid amount of \$49.84.
The DRS records also describe a transient timeout message that conflicts
with the persistent Paid state. The retained procedure calls for
reopening My Bookings and inspecting the orders to establish what was
actually saved. Seat-letter checks belong to the seat-selection views;
the order detail shown here supports the route and payment checks.

\paragraph{Reuse during the later booking task.}
After further practice, the final T065 actor reads the accumulated
booking experience and the GUI capture, booking-flow, and form-filling
playbooks in its first two iterations. Its eventual itinerary uses
G98 and D756, rather than simply copying the G102/G122 route discussed
in earlier memory. The final observation records two Paid orders for
March 25, 2026, a total of \$177.26, and arrival at Beijing South at
19:58. This illustrates retrieval of procedural knowledge alongside
fresh choices in the current interface. The recorded score changes
from 0 to 1, but the final run also receives an explicit current-date
clarification while the baseline books the year 2027. We therefore use
T065 primarily to illustrate memory acquisition and reuse; T049 and
T044 below provide the more direct artifact-level score comparisons.

\FloatBarrier
\subsection{Deep Recursive Self-exploration: From a Failed Repair to a Reusable Solution}
\label{app:case-presentation-learning}

T049 asks the agent to repair a GoogleNet presentation using the
accompanying paper as a reference. Slide 2 has displaced text and card
elements, and slides 3 and 4 contain misarranged Inception-module boxes
and arrows. The repair must preserve non-arrow dimensions, font sizes,
and appearance while restoring the figures. By the start of DRS,
the actor already has the 89,086-byte corpus acquired through eight
exploratory projects. Target practice now exposes a gap in how those
skills are applied to the actual presentation.

\begin{figure}[p]
  \centering
  \includegraphics[width=0.93\linewidth]{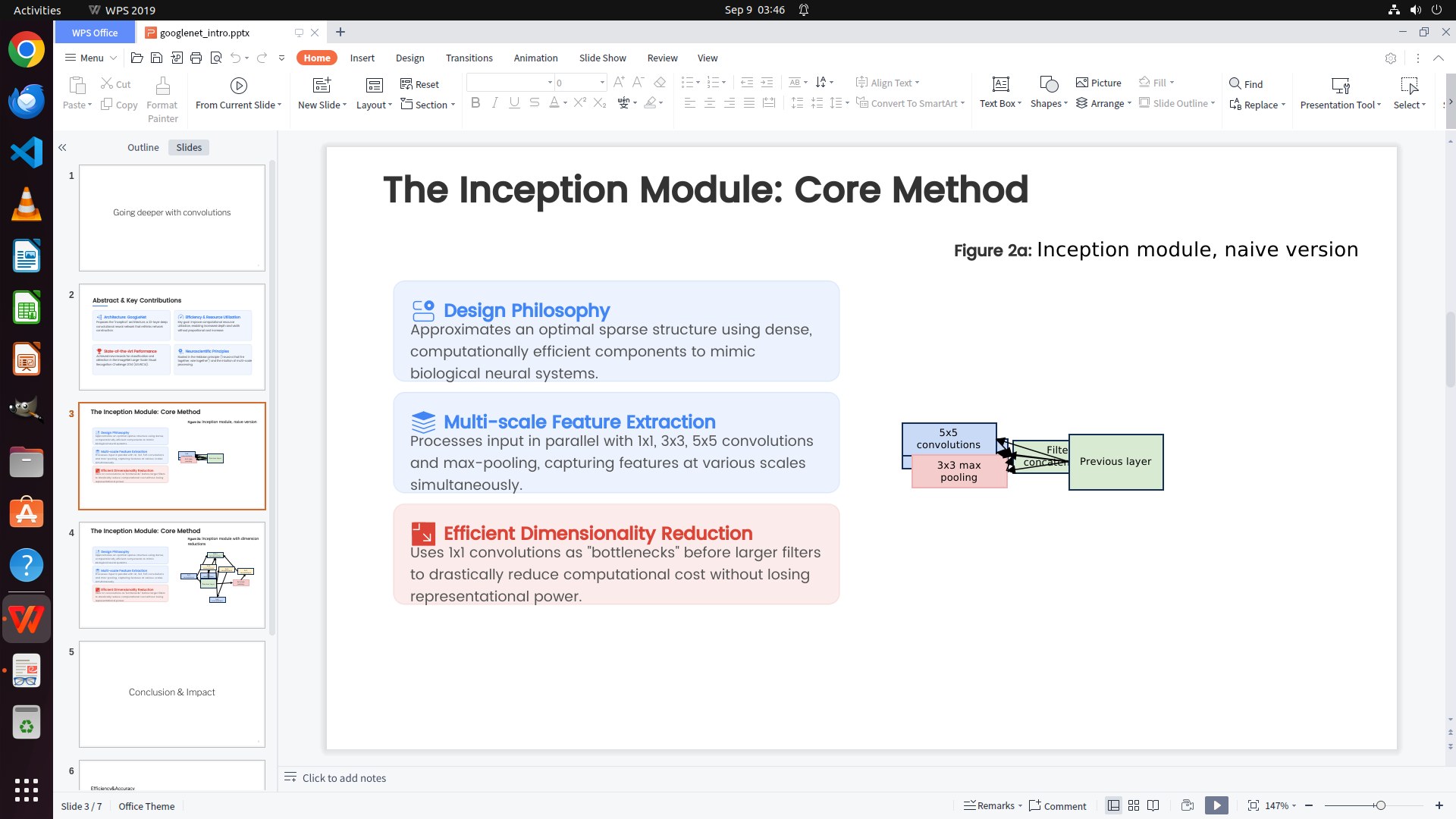}
  \par\smallskip\small (a) DRS, target cycle 1: internal FAIL; boxes remain piled.
  \par\medskip
  \includegraphics[width=0.93\linewidth]{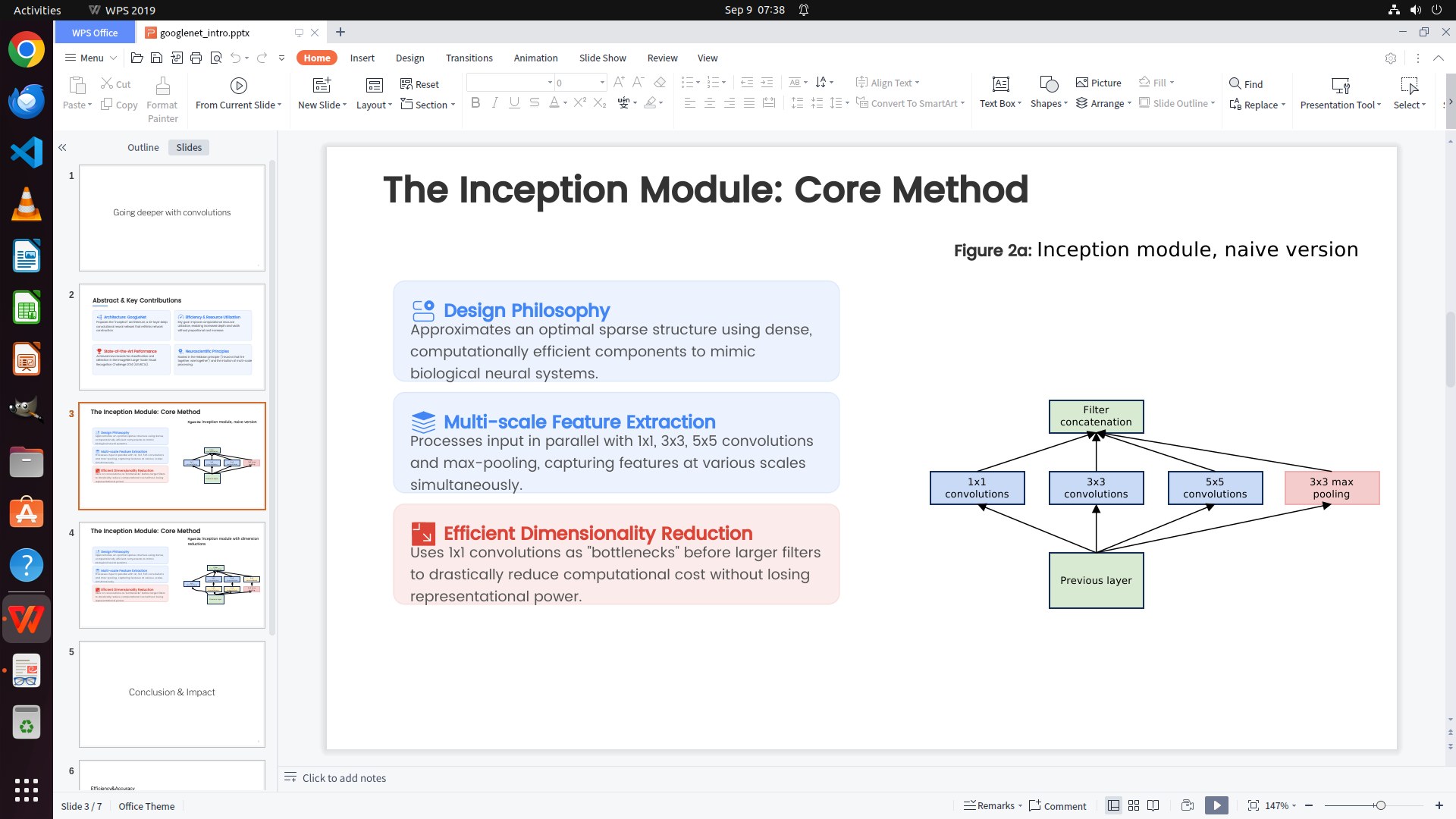}
  \par\smallskip\small (b) DRS, target cycle 2: internal PASS after memory updates and practice.
  \caption{The learning sequence within T049's DRS stage, shown in
  the original live WPS screenshots. The first attempt leaves the boxes
  at their corrupted positions. After recording the interpretation
  failure and completing a supplementary repair exercise, the next
  attempt reconstructs the reference layout. Both images belong to
  DRS practice; the separate memory-free baseline comparison appears
  in Figure~\ref{fig:case-t049-evaluation}.}
  \label{fig:case-t049-learning}
\end{figure}

\paragraph{First target cycle: precise operations under the wrong interpretation.}
The first attempt treats the restriction on element properties as a
prohibition on moving the figure boxes. It adjusts arrows while leaving
the corrupted box positions in place. The live WPS screenshot in
Figure~\ref{fig:case-t049-learning}(a) shows the result: the branch boxes
remain piled together, with arrows connecting an arrangement that does
not resemble the reference. The internal verifier rejects the candidate.

The memory update names the mistake explicitly:
\begin{quote}
  \small
  ``WHY FAIL (decisive): I FROZE THE FIGURE BOXES at their corrupted
  positions.''
\end{quote}
The same record explains the intended repair: move the boxes
\emph{position-only} into the paper's arrangement and then reconnect the
arrows. It preserves the competing readings of the task, the observed
layout defects, and the decoded reference geometry. Memory increases
to 110,546 bytes in fourteen files. The failure has supplied a concrete
decision rule for the next attempt: interpret the preservation
constraints together with the defect that the task asks to repair.

\paragraph{A supplementary branch tests the revised rule.}
Between the two target cycles, the curriculum introduces a related
synthetic presentation, \texttt{velvet\_signal\_path\_deck.pptx}.
This exercise also has piled boxes and damaged connectors, but includes
declared-correct elements that can anchor the repair. The actor applies
the newly recorded interpretation at its first render, fits a uniform
mapping from the reference PDF to the intact slide elements, validates
that mapping on an untouched connector, and then moves the displaced
boxes and repairs the arrows. The verifier returns PASS.

Together with the returning target attempt, this practice adds a useful
distinction to the memory: fit a mapping when
correct in-place anchors agree, and construct the layout directly from
the reference when the damaged figure has no usable anchors. The stored
procedure now describes when to use each method, as well as how to
calculate the resulting coordinates.

\paragraph{Second target cycle: a verified repair becomes a precedent.}
Returning to the GoogleNet deck, the actor checks its identity against
the retained failure record and uses the paper-derived arrangement.
It moves the boxes without changing their dimensions and connects the
arrows to the appropriate edge-center anchors. The live application
now displays a row of four branches between Previous layer and Filter
concatenation, as shown in Figure~\ref{fig:case-t049-learning}(b).
The internal verifier returns PASS, and consolidation adds the
successful repair record. At the final freeze, memory contains sixteen
files and 141,312 bytes.

The failure and success records remain available together. The failure
explains why an earlier choice was wrong; the success specifies the
working construction, the artifact identity checks, and the anchor
convention. In this case, DRS turns broad geometric
editing skills into a concrete, retrievable solution for the target
presentation.

\FloatBarrier
\subsection{Deep Recursive Self-exploration: Refining Video Edits Through Repeated Inspection}
\label{app:case-video-learning}

T044's DRS history adds seven supplementary projects and five target
outcomes to the editing knowledge acquired during BRS. The retained
images show how the agent checks the same operation at several levels:
the watermark region, the whole frame, recognizable content within the
frame, and the relationship between the saved project and its export.
These inspections enrich the existing editing and verification manuals,
which the final actor later reads before editing the evaluation video.

\begin{figure}[!htbp]
  \centering
  \begin{minipage}[t]{0.48\linewidth}
    \centering
    \includegraphics[width=\linewidth]{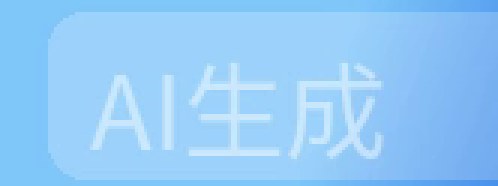}
    \par\smallskip\small (a) Enlarged source badge.
  \end{minipage}\hfill
  \begin{minipage}[t]{0.48\linewidth}
    \centering
    \includegraphics[width=\linewidth]{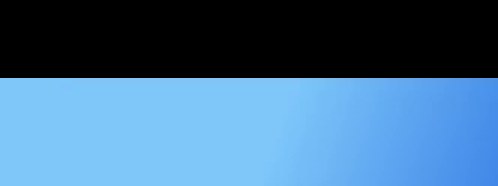}
    \par\smallskip\small (b) Processed region.
  \end{minipage}
  \caption{T044, supplementary DRS project 1: runtime close-ups used
  to inspect the watermark region. The original badge is visible in
  the source; the processed view shows the resulting black padding and
  retained image content.}
  \label{fig:case-drs-video-badge}
\end{figure}

\paragraph{A supplementary clip isolates the watermark boundary.}
Figure~\ref{fig:case-drs-video-badge} shows enlarged input and output
regions from the first supplementary project. The practice uses a
$640\times360$ clip with an AI-generated badge near the upper edge.
The actor removes a 68-pixel band and retains the remaining content
with symmetric 34-pixel padding. The verifier checks the project,
dimensions, output size, and the visible result, returning PASS.
This adds a small, directly inspected example of choosing the crop
from the actual badge boundary to the broader BRS experience.

\begin{figure}[!htbp]
  \centering
  \begin{minipage}[t]{0.43\linewidth}
    \centering
    \includegraphics[width=\linewidth]{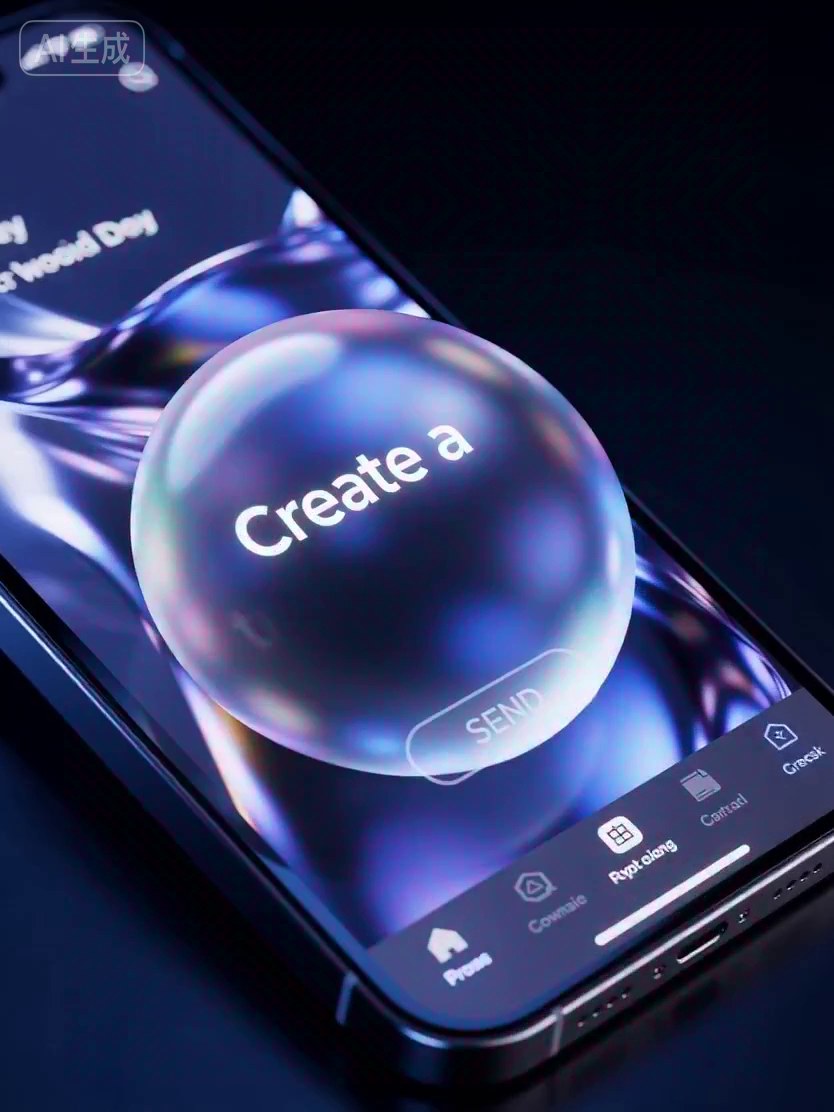}
    \par\smallskip\small (a) Source frame inspected in cycle 1.
  \end{minipage}\hspace{0.04\linewidth}
  \begin{minipage}[t]{0.43\linewidth}
    \centering
    \includegraphics[width=\linewidth]{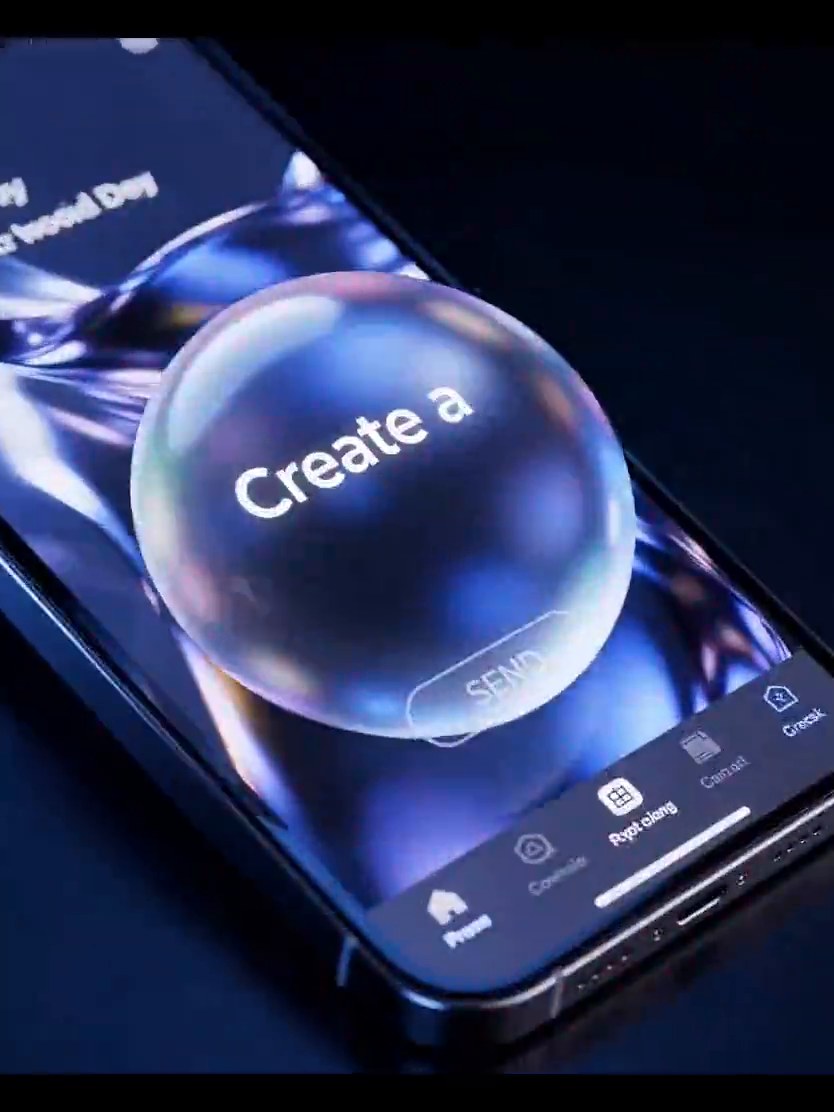}
    \par\smallskip\small (b) Exported frame from the same cycle.
  \end{minipage}
  \caption{T044, DRS target cycle 1: whole-frame inspection of the
  watermark-removal result. The phone scene makes the retained content
  easy to locate, while the frame boundaries reveal the added padding.
  Both images come from the same practice cycle.}
  \label{fig:case-drs-video-cycle-one}
\end{figure}

\paragraph{Whole frames establish the effect on the target video.}
The first target cycle inspects the portrait clip at a recognizable
phone scene (Figure~\ref{fig:case-drs-video-cycle-one}). The top-left
badge disappears in the export, and black bars appear at both ends of
the frame. The phone outline, spherical graphic, and application icons
provide visible landmarks for checking what the edit preserves.
The source and output are retained at the same $834\times1112$
resolution. This is an early target-practice observation that subsequent
cycles extend with more focused checks.

\begin{figure}[!htbp]
  \centering
  \includegraphics[width=0.94\linewidth]{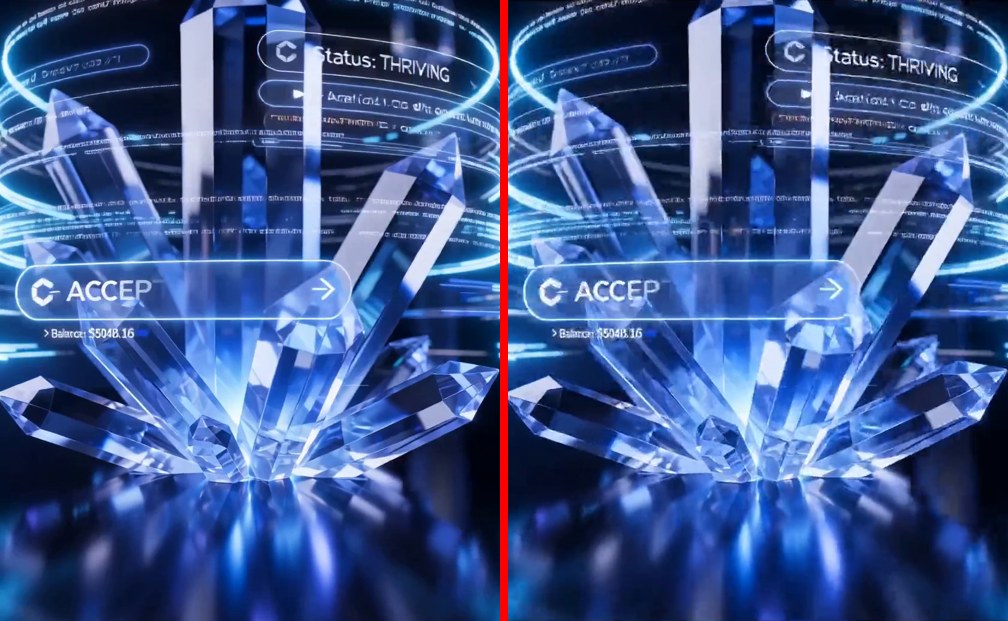}
  \caption{T044, DRS target cycle 2: the runtime comparison of
  corresponding content regions. The left view is the source after
  accounting for the top crop; the right is the exported content.
  The verifier uses the crystal contours and interface details to
  inspect geometric fidelity, alongside numerical image checks.}
  \label{fig:case-drs-video-content}
\end{figure}

\paragraph{Content alignment checks for distortion.}
Cycle 2 compares the exported content with the source after accounting
for the 76-pixel crop (Figure~\ref{fig:case-drs-video-content}).
This removes the expected framing difference from the comparison and
lets the verifier inspect the crystal edges and nearby interface
elements. The recorded inspection describes nearly identical geometry
with some compression softening; the accompanying analysis reports a
median structural-similarity score of approximately 0.99. The lesson
retained in the verification procedure is to compare corresponding
content regions when checking for scaling or stretching.

\begin{figure}[!htbp]
  \centering
  \includegraphics[width=0.91\linewidth]{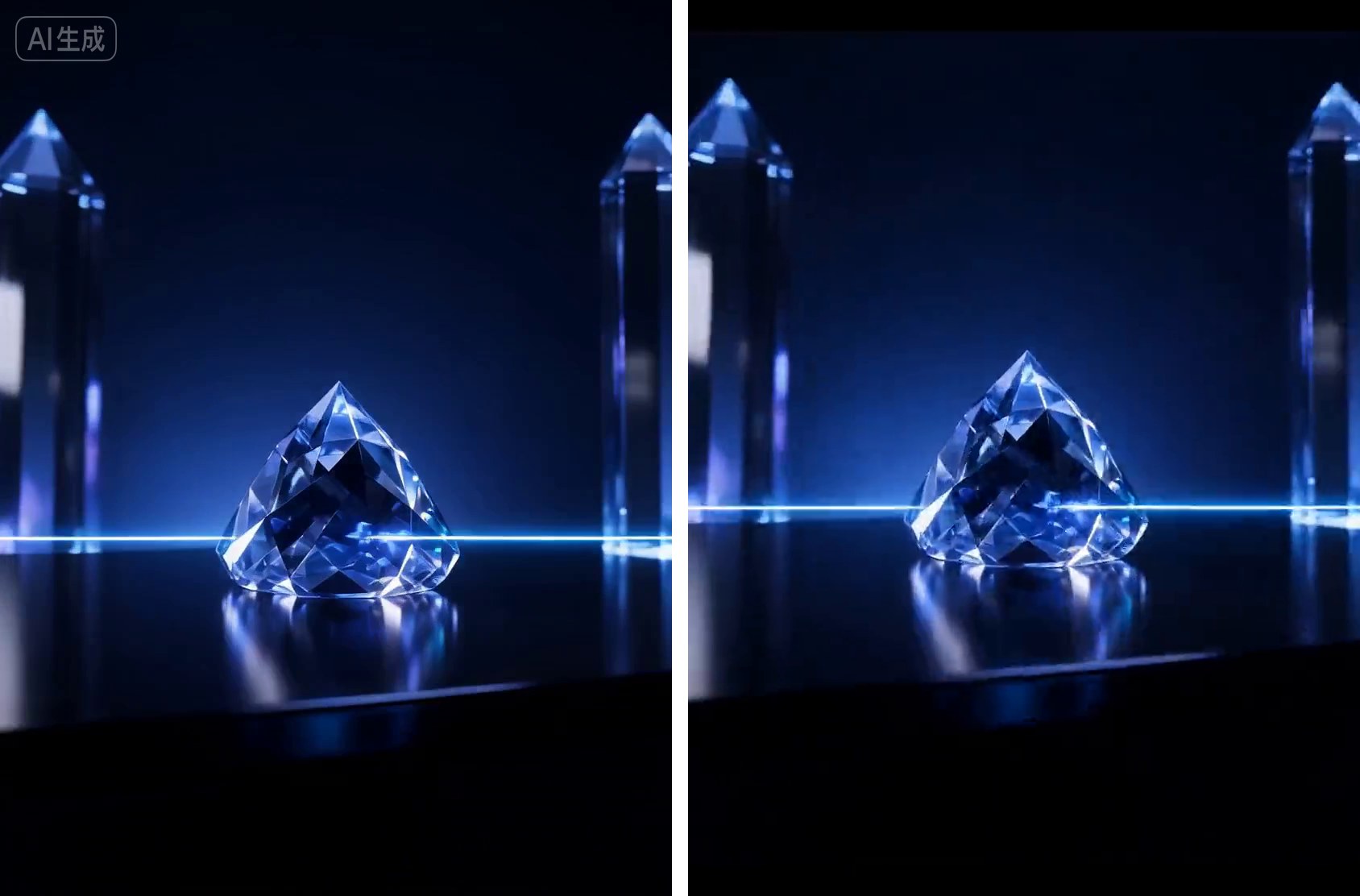}
  \caption{T044, DRS target cycle 4: a source/export comparison
  assembled during the original runtime inspection. The source is on
  the left and the export on the right. The later scene exposes the
  watermark boundary, padding, and retained crystal geometry together.}
  \label{fig:case-drs-video-cycle-four}
\end{figure}

\paragraph{Another scene checks the procedure across the clip.}
Cycle 4 retains a side-by-side comparison at a later crystal scene
(Figure~\ref{fig:case-drs-video-cycle-four}). The source watermark is
visible above the left crystal, while the corresponding exported frame
has a black top bar. The horizontal light line and crystal facets make
the upward displacement from cropping visible and provide another set
of landmarks for inspecting proportions. Sampling this different scene
extends the evidence beyond the phone frame used in the first cycle.

\begin{figure}[!htbp]
  \centering
  \includegraphics[width=0.96\linewidth]{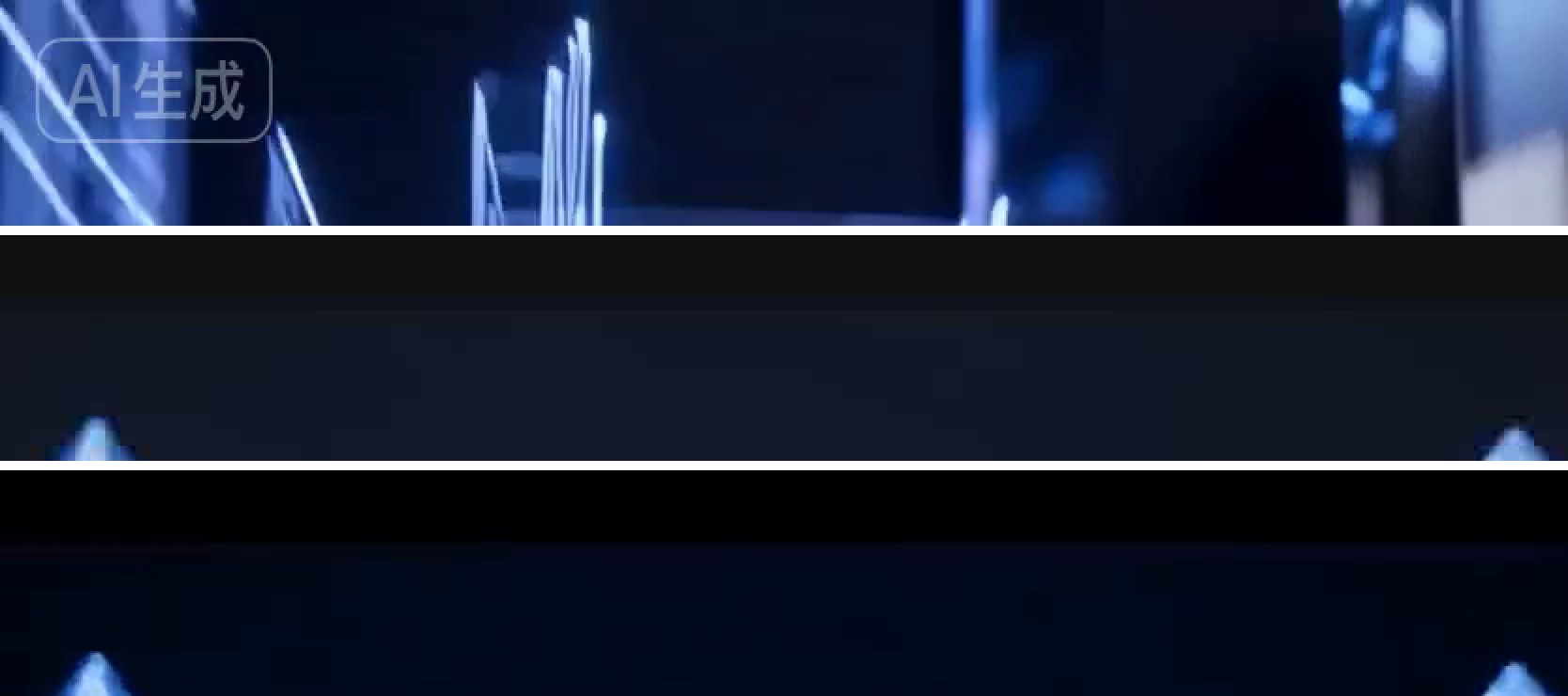}
  \caption{T044, DRS target cycle 4: the original three-way runtime
  inspection, with the source strip at the top, saved-project render
  in the middle, and exported strip at the bottom. The watermark is
  visible only in the source, and the export includes black padding.
  These views check both deliverables produced by the editing procedure.}
  \label{fig:case-drs-video-project-export}
\end{figure}

\paragraph{The saved project and export are both inspected.}
The same cycle also retains a three-way check of the top image region
(Figure~\ref{fig:case-drs-video-project-export}). The source contains
the badge; a render of the saved MLT project and the exported video
both remove it. Their framing differs in this inspection, with black
padding visible in the exported strip. Keeping both views makes the
relationship between the editable project and delivered video part of
the verification procedure. This is directly relevant to the final
task, which evaluates the native crop settings as well as the video.

\begin{figure}[!htbp]
  \centering
  \includegraphics[width=0.90\linewidth]{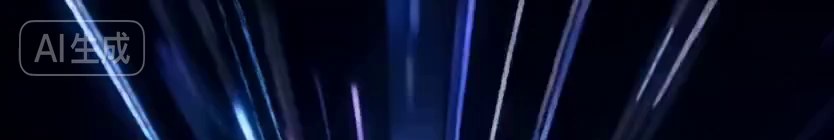}
  \par\smallskip\small (a) Source top strip: the badge remains visible.
  \par\medskip
  \includegraphics[width=0.90\linewidth]{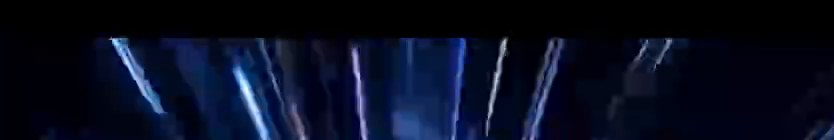}
  \par\smallskip\small (b) Processed top strip: padding and retained content.
  \par\medskip
  \includegraphics[width=0.48\linewidth]{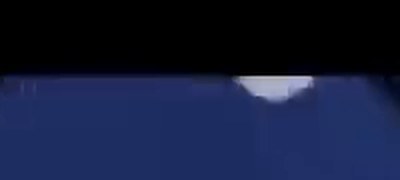}
  \par\smallskip\small (c) Additional enlarged export detail.
  \caption{T044, DRS target cycle 5: retained runtime views of the
  watermark boundary at two inspection scales. The input strip makes
  the badge location explicit; the output views support the final
  visual check that its pixels have been removed.}
  \label{fig:case-drs-video-final-strips}
\end{figure}

\paragraph{A final close-up rechecks the boundary.}
The fifth target cycle again inspects the original upper strip, the
processed strip, and an enlarged export detail
(Figure~\ref{fig:case-drs-video-final-strips}). These views concentrate
on the region where a shallow crop could leave badge pixels behind.
Across the DRS sequence, the memory therefore accumulates a layered
inspection method: measure the source boundary, check full-frame
composition, align content for distortion checks, and reopen both the
project and export. At the final freeze, these procedures and episode
records occupy 316,541 bytes in the same twelve-file corpus.

\FloatBarrier
\subsection{Without Memory Versus Frozen Memory: Which Errors Change the Score?}
\label{app:case-memory-comparison}

Table~\ref{tab:case-score-summary} summarizes the two principal
comparisons. We follow each score change down to the saved artifact
and the corresponding execution trace. These are selected historical
runs, with their recorded seeds and repair branches, rather than a
matched-seed memory ablation.

\begin{table}[!htbp]
  \centering
  \caption{Task scores in the principal memory-reuse case studies.
  Baseline uses no accumulated memory; the final run uses frozen memory
  after exploration and target practice. Scores are on a 0--1 scale.}
  \label{tab:case-score-summary}
  \small
  \begin{tabularx}{\linewidth}{@{}lrrrX@{}}
    \toprule
    Task & Baseline & Memory & Gain & Newly satisfied scoring checks \\
    \midrule
    T049 & 0.40 & 0.80 & +0.40 & Slide-3 arrows, completing its layout-and-arrow check \\
    T044 & 0.40 & 1.00 & +0.60 & Native crop height (+0.40) and Center-off setting (+0.20) \\
    \bottomrule
  \end{tabularx}
\end{table}

\paragraph{T049: retrieval changes the starting point of execution.}
The final actor reads both the GoogleNet failure and success records,
together with the memory index, in iteration 2. Its next message states
that the same corrupted deck has appeared before and that it should
verify the live file against the recorded MD5 and damaged coordinates.
After those checks, it reads the connector playbook, confirms the
reference geometry, and reuses the verified construction. The trace
therefore shows memory entering the repair before the main editing
decisions, followed by explicit checks that the remembered solution
applies to the current file.

The memory-free baseline also reconstructs the box layout by the end
of its run. The remaining difference is more specific than the dramatic
DRS failure shown earlier: connector 69 on slide 3, running from
Previous layer to the $5\times5$ convolution branch, is not anchored
accurately enough. Its start differs from the source box's top center
by 121,104 EMU horizontally and 145,345 EMU vertically. Both exceed the
scorer's per-axis tolerance of 100,000 EMU. In the final run with memory,
both errors are zero. The actor applies the recorded edge-center
convention and checks all nineteen arrows across the two figure slides.

\begin{figure}[!htbp]
  \centering
  \begin{minipage}[t]{0.49\linewidth}
    \centering
    \includegraphics[width=\linewidth,trim=520bp 138bp 20bp 180bp,clip]{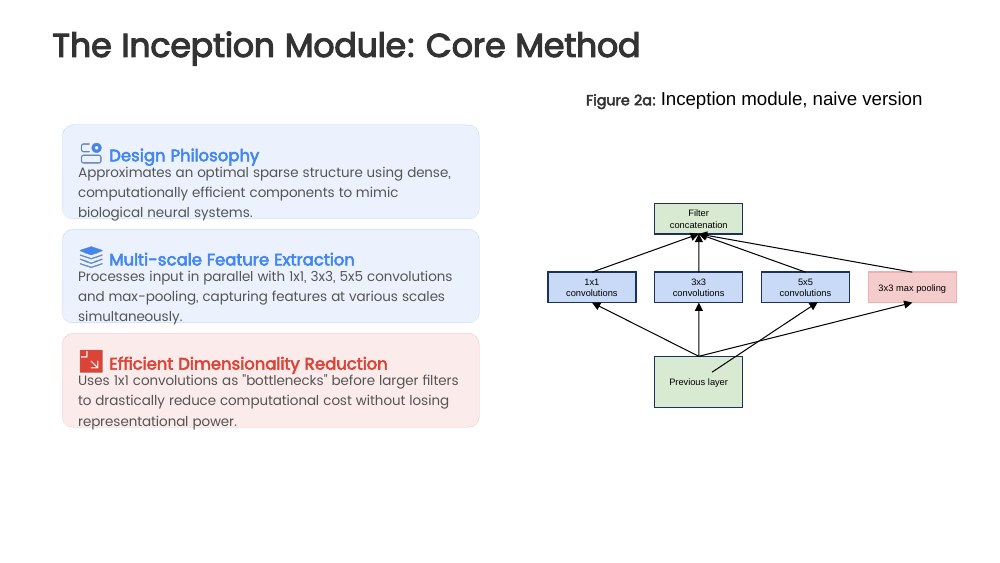}
    \par\smallskip\small (a) Without memory: task score 0.40.
  \end{minipage}\hfill
  \begin{minipage}[t]{0.49\linewidth}
    \centering
    \includegraphics[width=\linewidth,trim=595bp 119bp 5bp 245bp,clip]{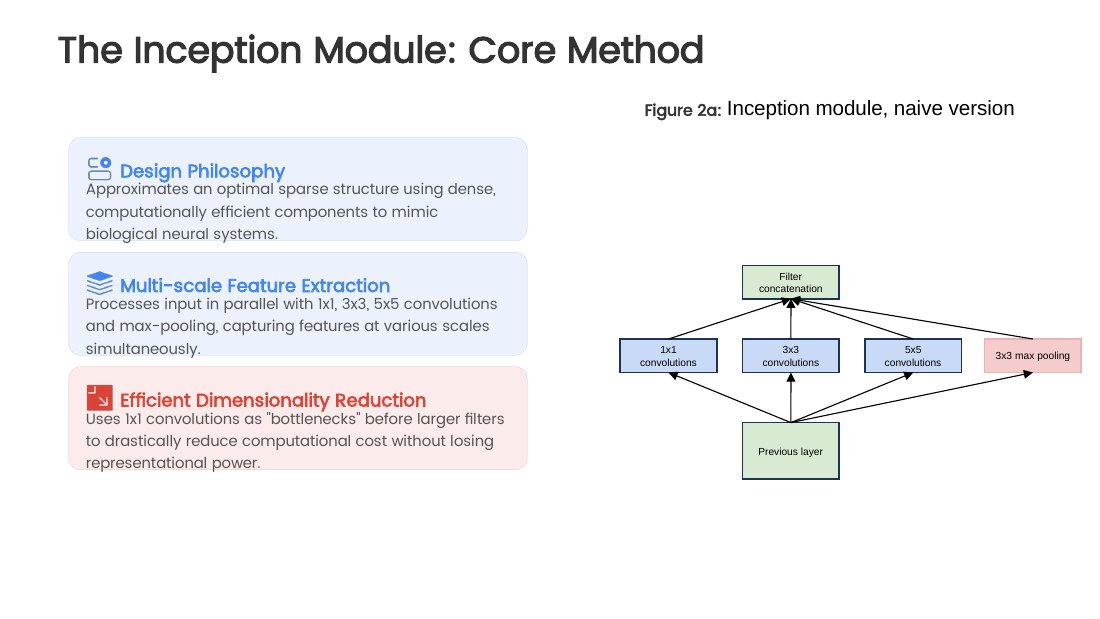}
    \par\smallskip\small (b) With frozen memory: task score 0.80.
  \end{minipage}
  \caption{Detail of T049's final slide-3 diagram in the two evaluation
  runs, cropped from artifact renders inspected by the verifier during
  execution. Both
  runs restore the broad arrangement; the decisive remaining baseline
  error is the start of connector 69, which misses the required
  Previous-layer anchor. The memory run places that endpoint exactly
  at the box's top center.}
  \label{fig:case-t049-evaluation}
\end{figure}

The saved scoring breakdown makes the consequence explicit
(Table~\ref{tab:case-t049-components}). Slide 3 requires both its layout
and arrows to pass. Correcting the endpoint changes that slide's
contribution from 0 to 0.40, accounting for the full task-level gain.
Slide 4 contributes 0.40 in both runs, and slide 2 contributes zero in
both, even though the internal verifier accepts the final repair.
Thus, the observed improvement is a completed figure-repair component,
with one presentation component still uncredited by the task scorer.

\begin{table}[!htbp]
  \centering
  \caption{T049 scoring breakdown from the retained artifact analysis.
  Replaying the scoring functions on the saved presentations reproduces
  the historical totals. EMU denotes the presentation's coordinate unit.}
  \label{tab:case-t049-components}
  \small
  \begin{tabularx}{\linewidth}{@{}Xrr@{}}
    \toprule
    Check & Without memory & Frozen memory \\
    \midrule
    Slide 2 contribution & 0.00 & 0.00 \\
    Slide 3 layout & Pass & Pass \\
    Slide 3 arrows & Fail & Pass \\
    Slide 3 contribution & 0.00 & 0.40 \\
    Slide 4 contribution & 0.40 & 0.40 \\
    \midrule
    Connector 69 start error, horizontal (EMU) & 121,104 & 0 \\
    Connector 69 start error, vertical (EMU) & 145,345 & 0 \\
    \midrule
    Total & 0.40 & 0.80 \\
    \bottomrule
  \end{tabularx}
\end{table}

\begin{figure}[!htbp]
  \centering
  \includegraphics[width=0.94\linewidth]{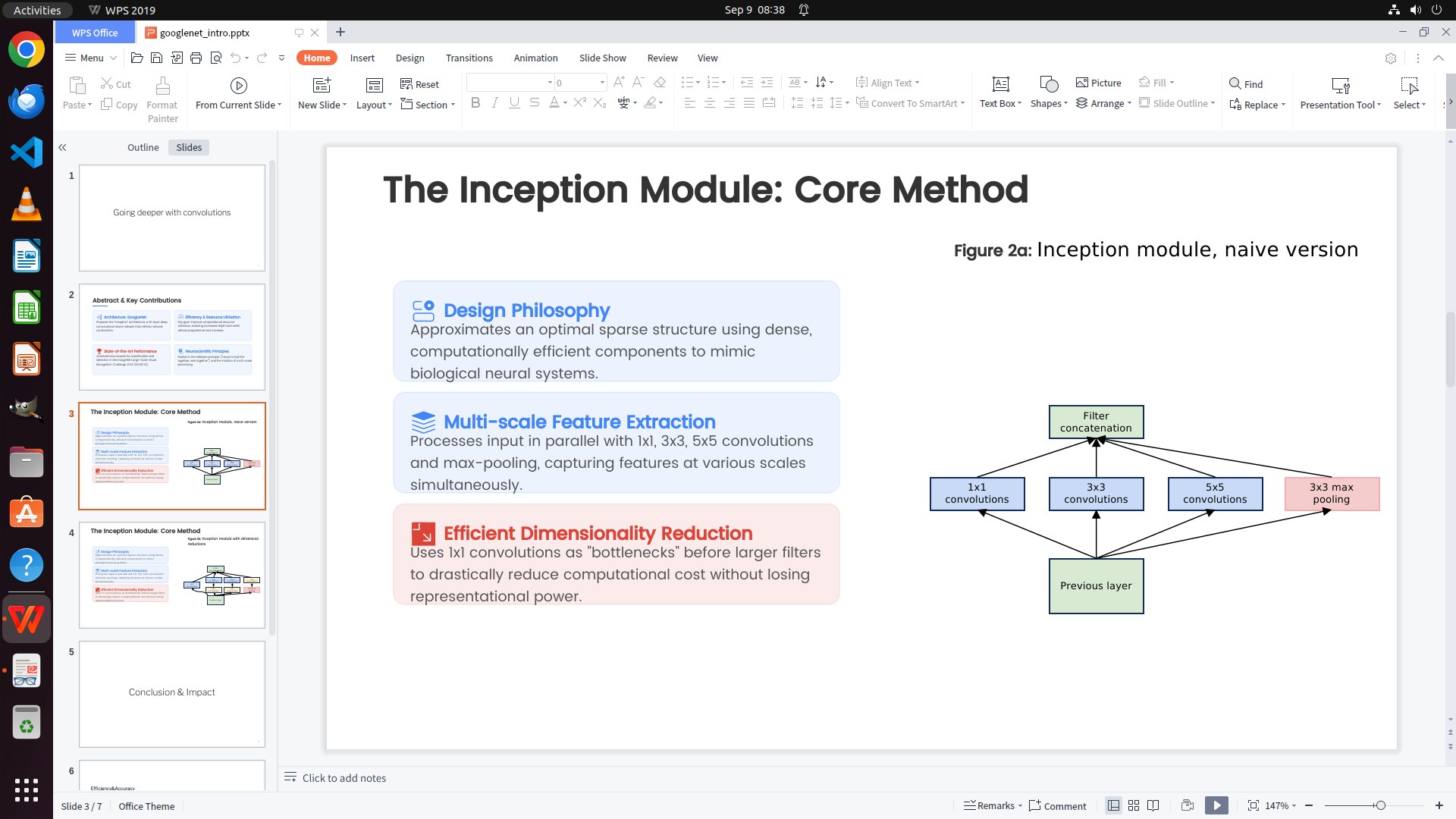}
  \caption{T049, test-time memory reuse: the repaired presentation in the live WPS
  window during final verifier inspection. This screenshot belongs to
  the frozen-memory evaluation whose task score is 0.80.}
  \label{fig:case-t049-live}
\end{figure}

The final repair is also visible in the live WPS application
(Figure~\ref{fig:case-t049-live}). In these two runs, actor iterations
decrease from 35 to 19 and recorded wall time decreases from 5,105.5 to
2,697.0 seconds. This is consistent with the trace's shift from deriving
a repair afresh to checking and executing a retained solution.

\FloatBarrier
\paragraph{T044: memory selects the application's native editing procedure.}
T044 asks for a video with the top watermark removed, symmetric
padding where needed, unchanged resolution, and no stretching or
zooming. It also requires both the saved Shotcut project and an exported
video no larger than 1 MiB. The baseline removes the top 100 pixels
using generic \texttt{avfilter.crop} and \texttt{avfilter.pad} filters,
with 50-pixel black bars above and below the remaining image. Its output
is 1,019,423 bytes at the original $834\times1112$ resolution.

The final actor with memory reads \texttt{shotcut-video-editing.md} in
iteration 4 and the native Shotcut operation notes in iteration 6.
These notes include the bundled rendering engine, crop-and-letterbox
procedures, export pitfalls, and output checks. The actor inspects the
watermark region and uses the GUI's \emph{Crop: Source} operation,
producing a project with \texttt{mlt\_service=crop},
\texttt{top=76}, and \texttt{center=0}. It exports through the bundled
\texttt{melt} engine. The final video retains the original resolution,
has 38-pixel symmetric bars, and is 986,760 bytes.

\begin{figure}[!htbp]
  \centering
  \begin{minipage}[t]{0.31\linewidth}
    \centering
    \includegraphics[width=\linewidth]{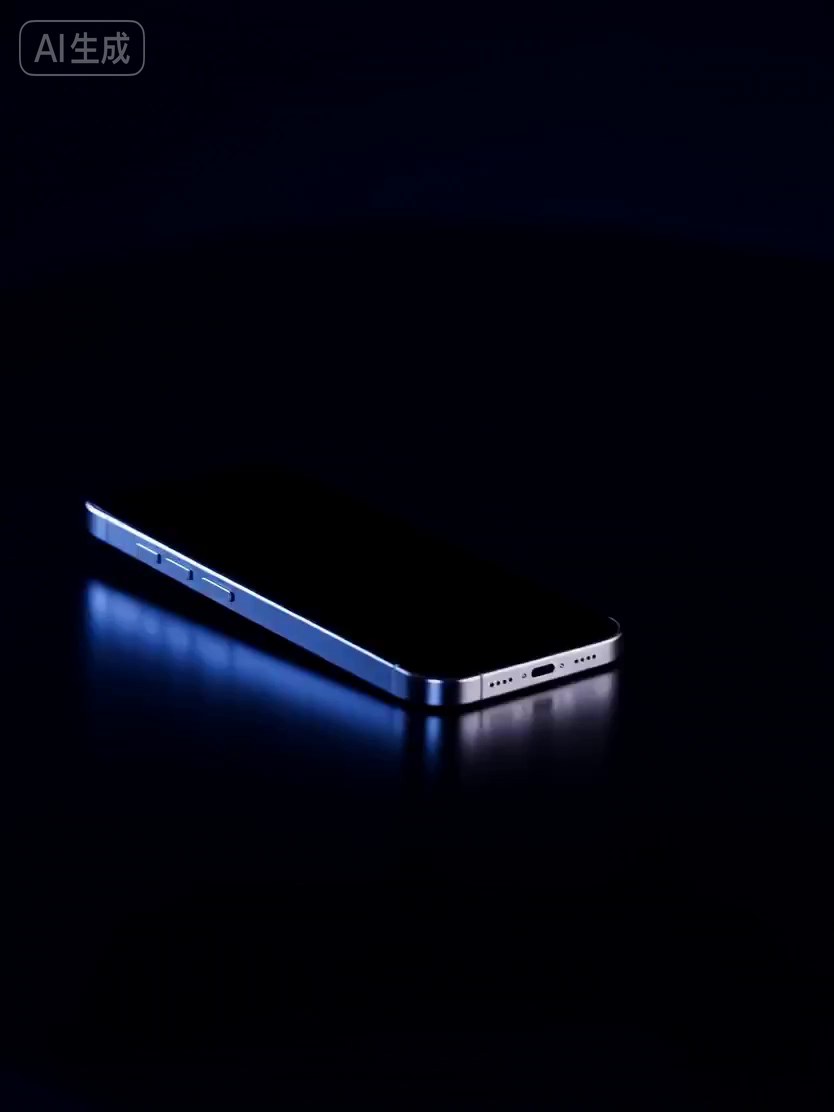}
    \par\smallskip\small (a) Source: top watermark.
  \end{minipage}\hfill
  \begin{minipage}[t]{0.31\linewidth}
    \centering
    \includegraphics[width=\linewidth]{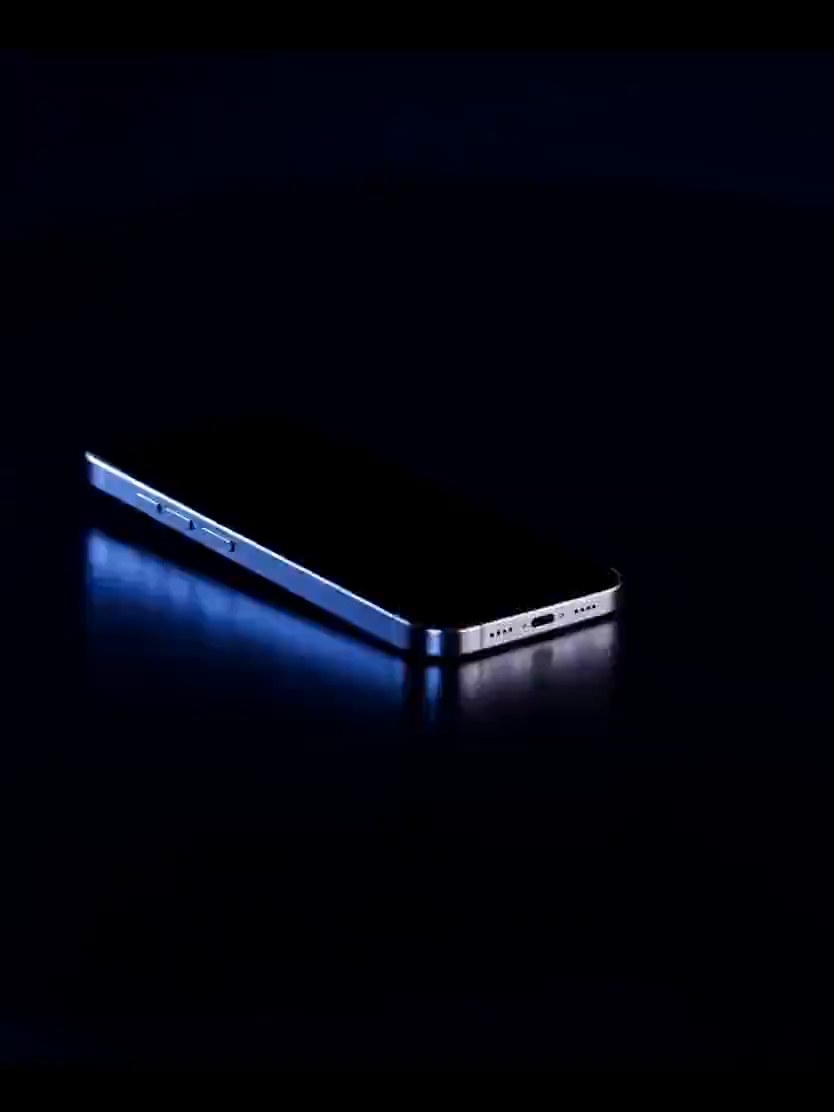}
    \par\smallskip\small (b) Baseline: 100-pixel crop.
  \end{minipage}\hfill
  \begin{minipage}[t]{0.31\linewidth}
    \centering
    \includegraphics[width=\linewidth]{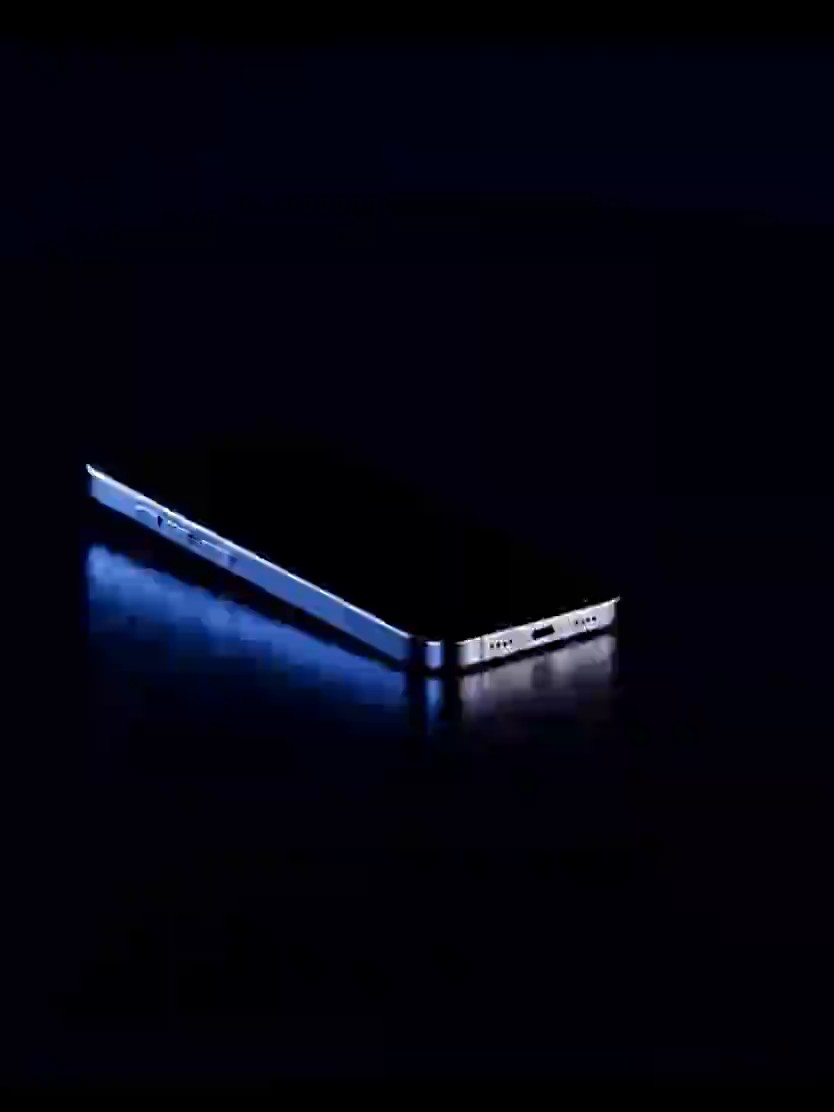}
    \par\smallskip\small (c) Memory: 76-pixel crop.
  \end{minipage}
  \caption{Video frames inspected during the original T044 runs.
  Panels (a) and (c) are the source/output views from the memory run;
  panel (b) is the baseline verifier's first-frame output view.
  These retained runtime frame views show watermark removal and
  symmetric padding in both outputs. The finer crop and the native
  project filter distinguish the memory run.}
  \label{fig:case-t044-runtime}
\end{figure}

\paragraph{Why this changes the score from 0.40 to 1.00.}
The task scorer inspects both the video and its project structure. It
awards 0.20 for the size cap and 0.20 for unchanged resolution, which
both outputs satisfy. Its remaining checks search for a native
\texttt{crop} or \texttt{qtcrop} filter: an accepted top crop between
75 and 85 pixels earns 0.40, and the Center-off condition earns 0.20.
The baseline's generic filters do not enter those checks. The memory
run's native filter is recognized, its 76-pixel crop falls in the
accepted range, and Center is disabled, so it receives all four
components.

The visual comparison and project analysis tell a complementary story.
Both runs remove the watermark and preserve the image proportions;
the improvement is a more precise crop expressed through the native
Shotcut workflow. Memory supplies application-specific knowledge about
which operation to use and how to validate its output. This case adds
a fully credited task, whereas T049 adds credit for a previously failed
slide. T044 also takes more actor iterations with memory, 42 versus 26,
so its gain is in task completion rather than shorter execution.

\FloatBarrier
\subsection{Reconstructing a Mechanical Part from Engineering Drawings}
\label{app:case-cad}

\begin{figure}[!htbp]
  \centering
  \includegraphics[width=0.94\linewidth,height=2.45in,keepaspectratio]{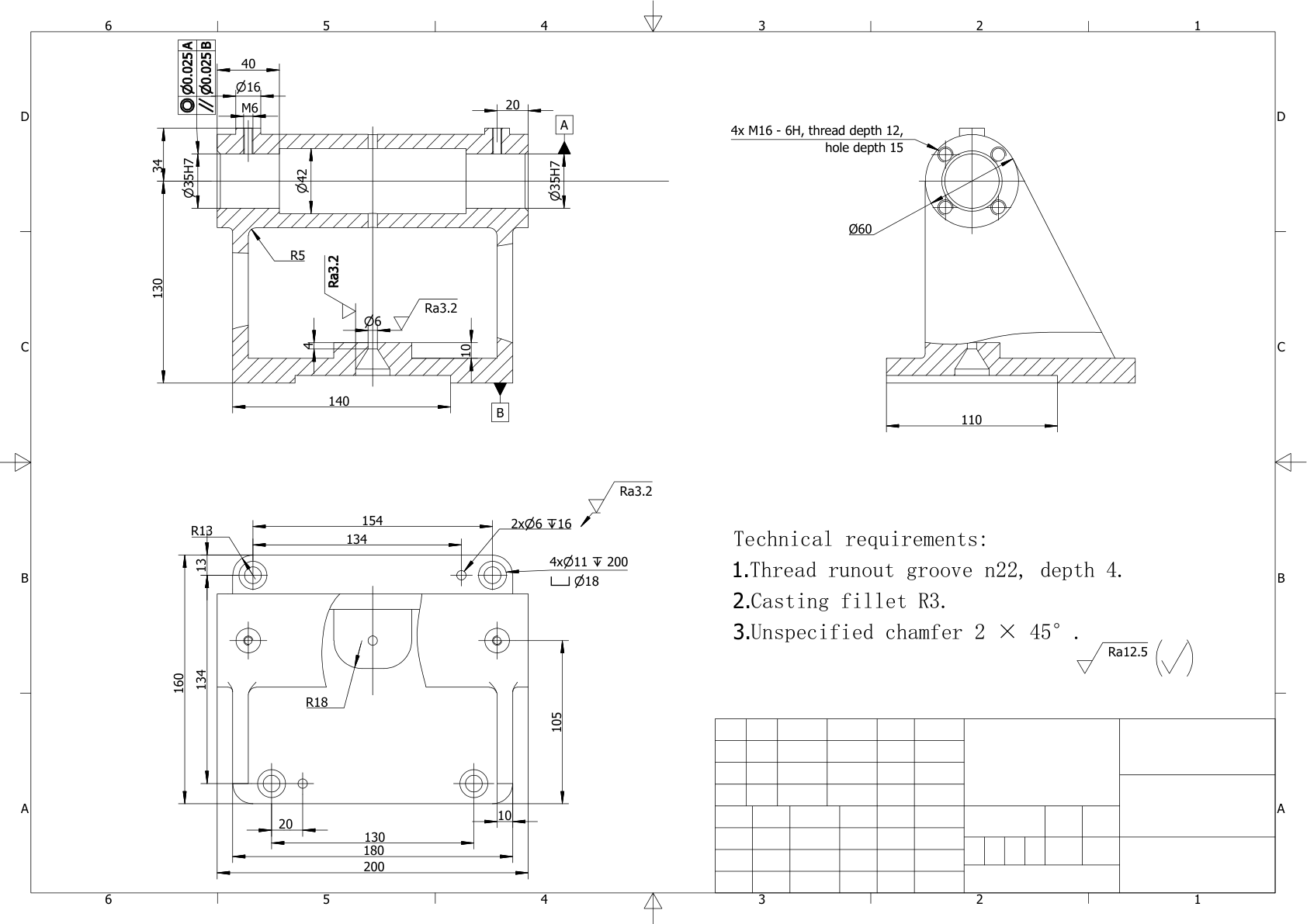}
  \par\smallskip\small (a) Supplied engineering drawing, rendered upright.
  \par\medskip
  \begin{minipage}[t]{0.48\linewidth}
    \centering
    \includegraphics[width=\linewidth,height=2.15in,keepaspectratio]{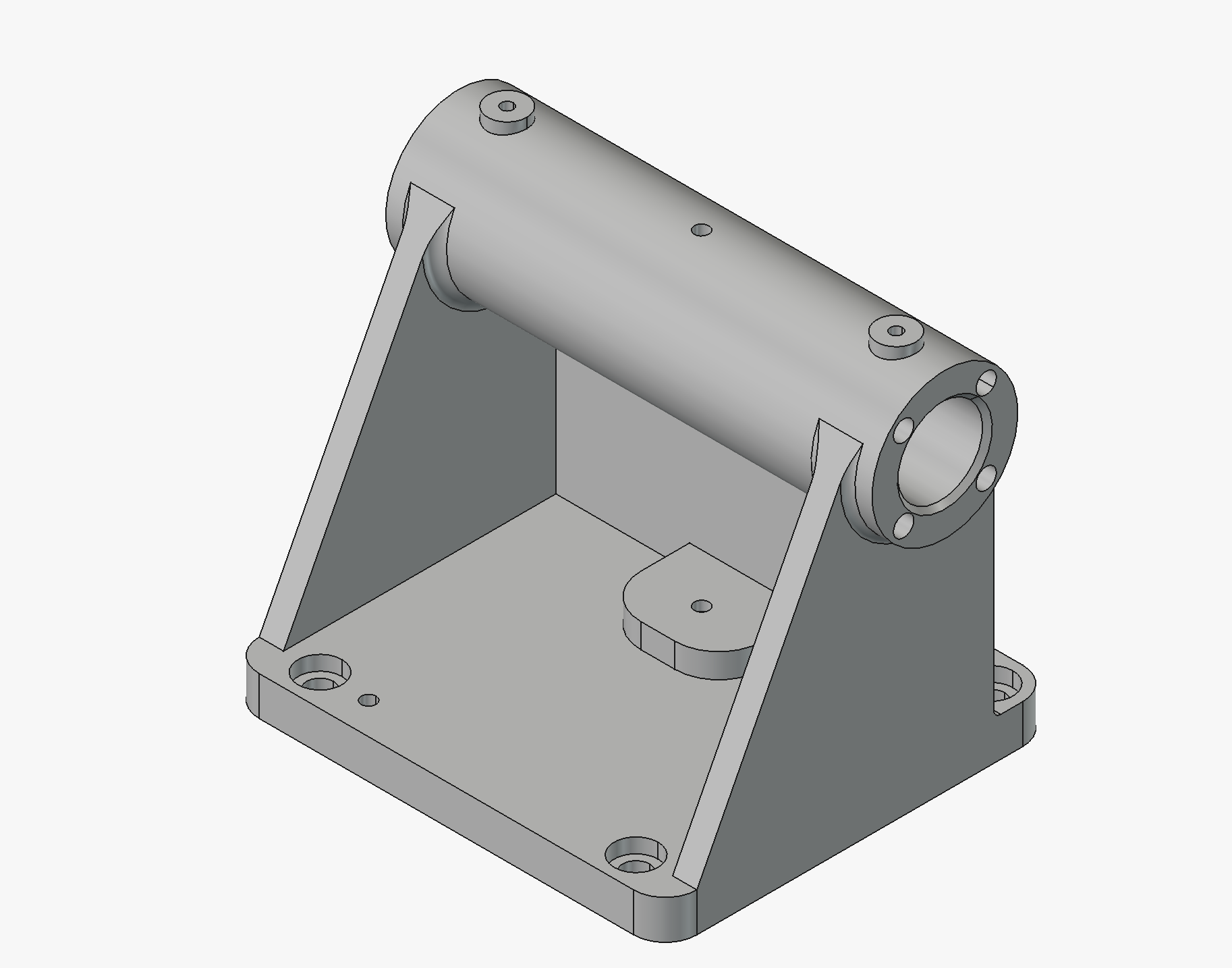}
    \par\smallskip\small (b) Supplied reference part.
  \end{minipage}\hfill
  \begin{minipage}[t]{0.48\linewidth}
    \centering
    \includegraphics[width=\linewidth,height=2.15in,keepaspectratio]{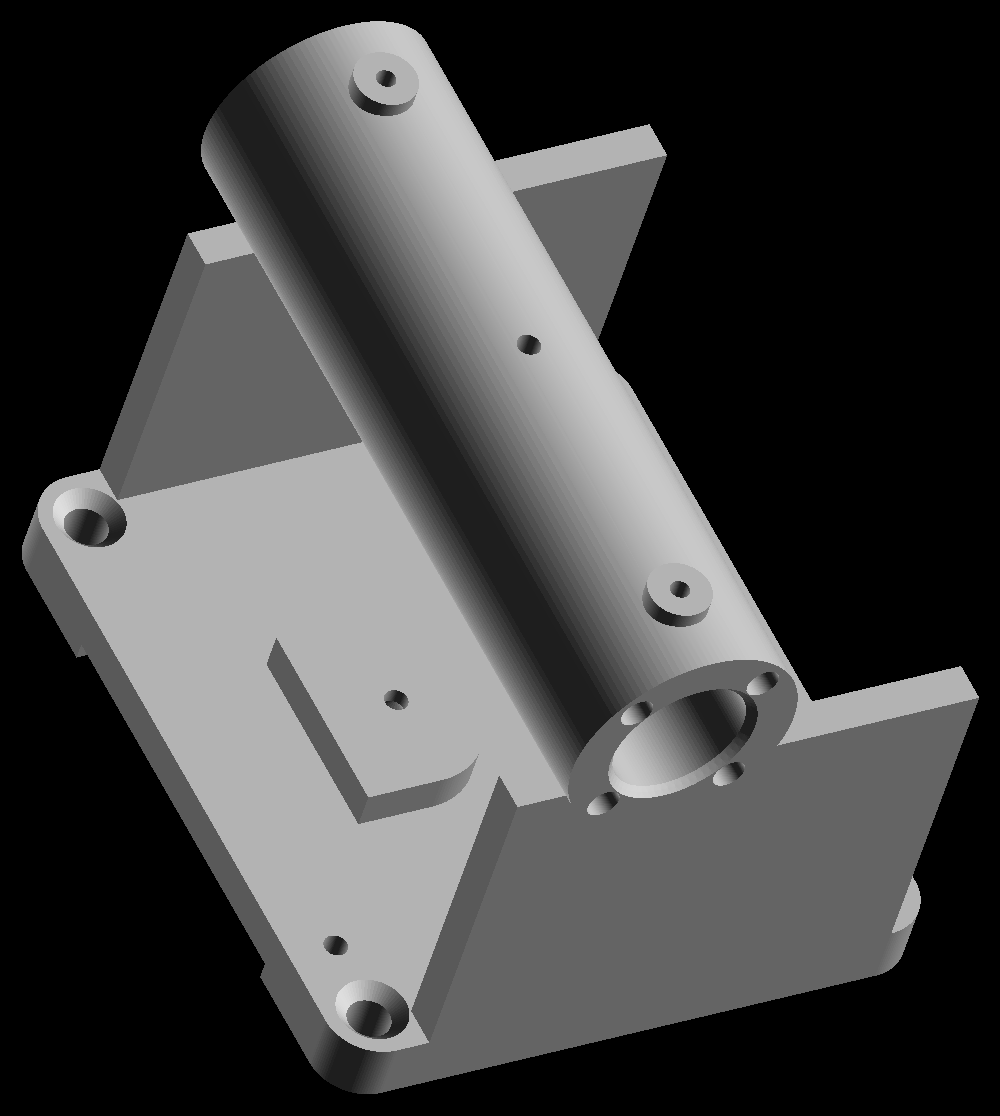}
    \par\smallskip\small (c) First evaluation: retained candidate render.
  \end{minipage}
  \caption{T103: from an engineering drawing to a programmatically
  constructed CAD model. Panel (c) is an original runtime render from
  the first RSI evaluation (score 0.6789), not the reference or the
  second evaluation. The rectangular supports and conical mounting-hole
  mouths remain visibly different from the reference.}
  \label{fig:case-cad}
\end{figure}

\paragraph{CAD reconstruction requires both interpretation and execution.}
T103 asks the agent to recreate a support bracket in FreeCAD from
\texttt{drawing.pdf} and a reference image, then export
\texttt{support\_bracket.step}. The part combines a bearing housing,
stepped bore, supporting webs, mounting holes, and a recessed base.
Completing it requires the agent to reconcile orthographic sections,
dimension annotations, and visible features before expressing their
relationships as a solid model. Figure~\ref{fig:case-cad} shows the
supplied inputs and a retained render from the first frozen-memory
evaluation.

\paragraph{Broad Recursive Self-exploration builds complementary CAD skills.}
Eight projects, executed in two waves of four, cover environment setup
and seven CAD exercises. A scanned-drawing project requires a gusseted
bracket with counterbored holes; an angled-brace project introduces
non-orthogonal construction; and a cover-plate project places a window
whose dimensions must be recovered from a reference-image grid.
These projects accumulate procedures for distinguishing scanned from
vector PDFs, interpreting drawing notes, constructing solids through
code, and checking a STEP export after re-import. All eight receive
internal PASS judgments, leaving 77,800 bytes of memory in eight files.
The shared knowledge therefore covers several prerequisites of the
target, rather than a single completed bracket.

\paragraph{Deep Recursive Self-exploration refines the interpretation rules.}
The first target attempt receives an internal FAIL amid disagreement
about the drawing's measurement frame. The curriculum agent then
proposes three flange-housing practices, whose outcomes are
FAIL, PASS, and PASS. Their retained records distinguish printed labels,
PDF coordinates, and physical dimensions, including a variant whose
labels are explicitly expressed in 96-dpi pixel units. A second target
attempt receives internal PASS and adds further procedures: anchor
dimension labels to arrowheads and extension lines, and combine
orthogonal sections to recover the recessed region, groove, taper, and
small passage. Every target and practice outcome contributes a memory
update. The resulting frozen corpus contains 147,415 bytes in the same
eight files. This historical run uses verifier-pass stopping for DRS.

\paragraph{Frozen memory supplies methods used in the final reconstruction.}
The evaluation traces show retrieval followed by application. In the
first evaluation, the actor agent reads the PDF and FreeCAD playbooks,
adopts their input-classification and execution rules, and later
explicitly reuses the earlier target's point-membership probes to check
material and void regions in the exported solid. The second evaluation
reads the memory index and the drawing-extraction, execution, FreeCAD,
and verification playbooks within its first five turns. It then derives
geometry from the current drawing and constructs the part through the
FreeCAD Part API. Thus, the retained experience provides both an
interpretation procedure and executable ways to check the resulting
artifact.

\paragraph{The score gain reflects improved, but incomplete, reconstruction.}
The archived memory-free baseline scores 0.2500; the two frozen-memory
evaluations score 0.6789 and 0.6897, respectively. The baseline already exports
a valid solid, so the improvement is not simply successful file creation.
The first RSI construction adds a cylindrical groove and a more detailed
central pad, while retaining incorrect support and mounting-hole
geometry visible in Figure~\ref{fig:case-cad}. These records demonstrate
reuse of task-relevant experience alongside higher partial credit on the
known target. The comparison uses an archival baseline rather than a
matched memory ablation, and the saved scores do not resolve the
contribution of individual features.

\FloatBarrier
\subsection{Producing a Radio Bumper in REAPER}
\label{app:case-reaper}

\begin{figure}[!htbp]
  \centering
  \includegraphics[width=\linewidth]{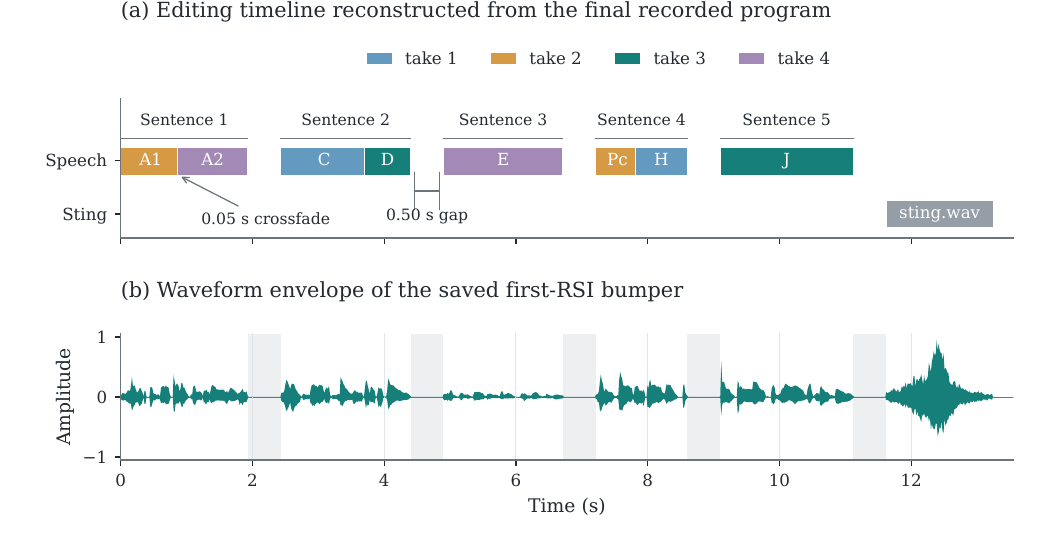}
  \caption{T085 radio-bumper assembly. (a) Source-take layout reconstructed
  from the final recorded construction program. (b) Waveform envelope of
  the saved first-RSI bumper; shaded bands mark the 0.50-second gaps.}
  \label{fig:case-t085-assembly}
\end{figure}

\paragraph{Audio production couples content selection with precise timing.}
\href{https://www.reaper.fm/}{REAPER} is a digital audio workstation
for arranging, processing, and rendering audio. T085 asks the agent to
assemble a radio bumper from recorded takes according to a supplied
fragment plan. The output must preserve the specified sources and order,
use approximately 0.05-second crossfades, leave 0.50-second gaps between
sentences, and place a closing sting 0.50 seconds after the final sentence.
That sentence must also be exported separately with its pitch raised by
two semitones and its duration halved. The task therefore requires both
interpreting which source fragments belong together and expressing that
interpretation through the application's editing and rendering controls.
Figure~\ref{fig:case-t085-assembly} connects the final source-fragment
arrangement to the audio delivered by the first frozen-memory evaluation.

\paragraph{Broad Recursive Self-exploration acquires reusable audio procedures.}
In this run, seven BRS projects cover source selection, precise splicing,
silence measurement, pitch and duration transformations, and complete
bumper construction. They leave 13 memory files containing 109,720 bytes.
The retained procedures include REAPER project construction and
command-line rendering, audio-content comparisons, and checks on the
exported waveform. This supplies the actor agent with practical methods
for constructing and inspecting audio artifacts before attempting T085.

\paragraph{Deep Recursive Self-exploration resolves ambiguity and revises rendering advice.}
The first target attempt receives an internal PASS and adds
\texttt{plan\_resolution\_pattern.md}. This record describes how to
combine phrase detection, cross-take comparisons, transcription, and
sentence grouping to resolve the terse fragment plan. It preserves the
chosen source spans together with qualifications about ambiguous choices.
The curriculum agent then selects six further practices. The first two
compare native item-and-fade assembly against preassembled sentence
sources, then repeat the comparison with tightly trimmed stereo audio.
The second practice exposes a mismatch between the rendered audio
samples; changing the saved resampling configuration to
\texttt{RENDER\_RESAMPLE 0 0 0} resolves it in the tested builds.
Memory is revised to qualify earlier renderer explanations. Four
additional practices examine analogous source-selection and ordering
problems in image composition. After a second target PASS and a
curriculum-ready decision, memory is frozen at 17 files and 227,605 bytes.
Both exploration stages record zero official evaluator calls.

\begin{figure}[!htbp]
  \centering
  \includegraphics[width=\linewidth]{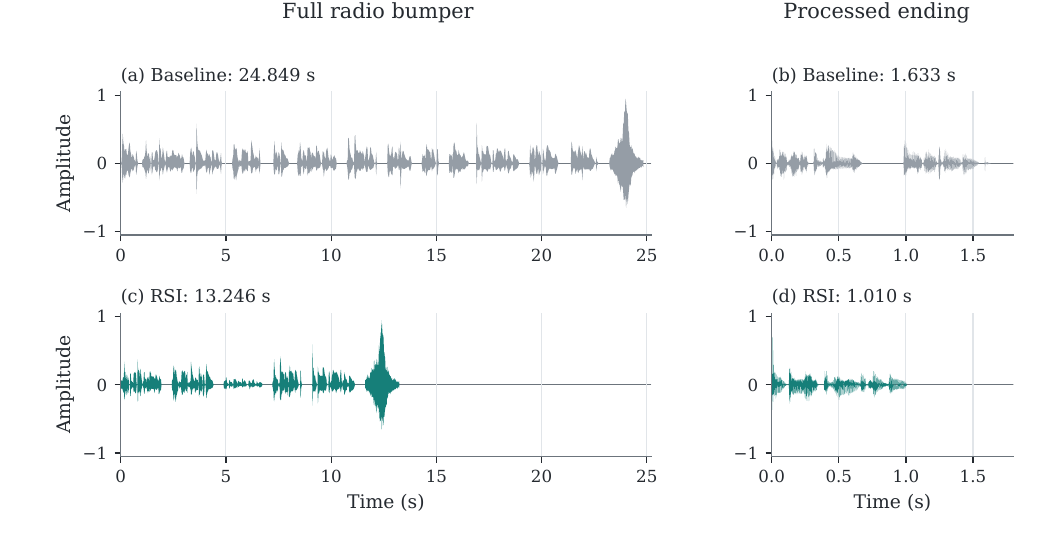}
  \caption{Waveform envelopes of the archived T085 exports: baseline
  (top) and the first frozen-memory RSI evaluation (bottom). Each column
  uses shared time and amplitude scales; panel titles give file durations.}
  \label{fig:case-t085-exports}
\end{figure}

\paragraph{Test-time execution reuses the retained interpretation and procedures.}
The first frozen-memory evaluation explicitly retrieves the full
plan-resolution record in iteration 15. Its construction program later
uses all eight source spans recorded in that memory and adopts the
revised resampling setting when building native REAPER projects.
The actor agent checks the emitted audio through transcription,
source comparisons, timing measurements, and paired pitch estimates.
The saved re-render check reports identical audio-sample payloads for
both deliverables. The trace thus connects a retained lesson to a
specific construction choice and a subsequent output check.
Figure~\ref{fig:case-t085-exports} shows the retained baseline and RSI
exports for both the complete bumper and the separately processed ending.

\paragraph{The score improvement is concentrated in sentence gaps and the processed ending.}
The archived baseline scores 0.6800, while the two frozen-memory
evaluations score 0.9417 and 0.9413, averaging 0.9415 as reported in
Section~\ref{sec:ablations}. Both evaluations reuse the same frozen
memory with host writeback disabled. Table~\ref{tab:case-t085-components}
shows the first evaluation's component scores: sentence-gap credit
increases from 0.8250 to 1.0000, and processed-final-sentence credit
increases from 0.2837 to 0.8996. The latter also removes the baseline's
binding 0.68 score cap; source-order credit remains unchanged.
The evaluator approximates sentence boundaries through acoustic activity,
so these components measure its recorded criteria rather than complete
semantic or perceptual correctness. This historical comparison documents
memory reuse on the known target, not a matched-budget estimate of each
practice's contribution.

\begin{table}[!htbp]
  \centering
  \caption{T085 component scores and overall partial score on a 0--1
  scale. RSI uses the first frozen-memory evaluation; the overall score
  includes the benchmark's weights and caps.}
  \label{tab:case-t085-components}
  \small
  \begin{tabularx}{\linewidth}{@{}Xrr@{}}
    \toprule
    Component & Baseline & RSI (draw 1) \\
    \midrule
    Source order & 0.9062 & 0.9062 \\
    Sentence gaps & 0.8250 & 1.0000 \\
    Closing sting & 1.0000 & 1.0000 \\
    Processed final sentence & 0.2837 & 0.8996 \\
    Basic duration & 1.0000 & 1.0000 \\
    \midrule
    Overall partial score & 0.6800 & 0.9417 \\
    \bottomrule
  \end{tabularx}
\end{table}

\end{document}